\documentclass[twocolumn]{fairmeta}
\usepackage{lmodern}

\usepackage{colortbl}
\usepackage{fontawesome5}

\usepackage{amsmath}
\usepackage{amssymb}
\usepackage{xspace}
\usepackage{needspace}
\usepackage{etoc}
\AtBeginDocument{\etocdepthtag.toc{main}}
\usepackage[symbol]{footmisc}
\usepackage{enumitem}
\usepackage{pgf}
\usepackage{pgffor}
\usepackage{wrapfig}
\usepackage{amsmath}
\newcommand{\themodel}{\textsc{Scale4D}\xspace}

\newcommand{\papertitle}{Metric 4D Modeling for Embodied Perception}

\newif\ifmetareport

\newif\ifflywheelmock
\flywheelmocktrue

\definecolor{tabfirst}{HTML}{FFCCCC}
\definecolor{tabsecond}{HTML}{FFE6B3}
\definecolor{tabthird}{HTML}{FFFFCC}

\newcommand{\first}{\cellcolor{tabfirst}}
\newcommand{\second}{\cellcolor{tabsecond}}
\newcommand{\third}{\cellcolor{tabthird}}

\definecolor{metabluelight}{HTML}{CCE0F8}
\definecolor{metabluemedium}{HTML}{6D92BF}

\renewcommand{\dbltopfraction}{0.9}

\metareporttrue
\usepackage{float}
\let\cite\citep
\setcitestyle{numbers,square,comma}
\makeatletter\def\NAT@sort{\@ne}\def\NAT@cmprs{\@ne}\makeatother

\title{\papertitle}

\renewcommand{\authorformat}[2][]{{\sffamily\bfseries\mbox{#2$^{#1}$}}}
\renewcommand{\author}[2][]{\addtolist[#1]{#2}{\authorlist}{\authorformat}{,\penalty0\hspace{0.15em}}}
\DeclareRobustCommand{\metamark}{\raisebox{0.05em}{\includegraphics[height=0.45em,trim=0 0 2590 0,clip]{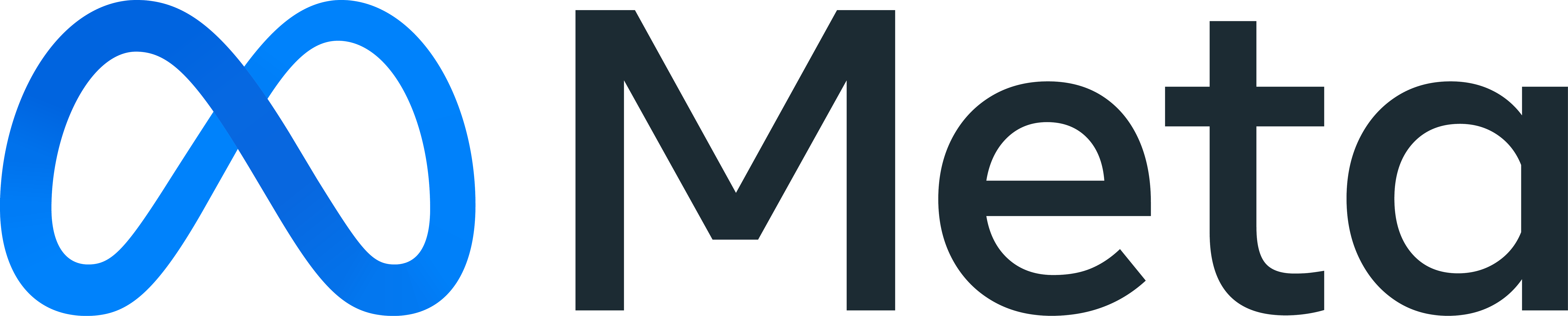}}}
\DeclareRobustCommand{\uvamark}{\raisebox{0.05em}{\includegraphics[height=0.55em]{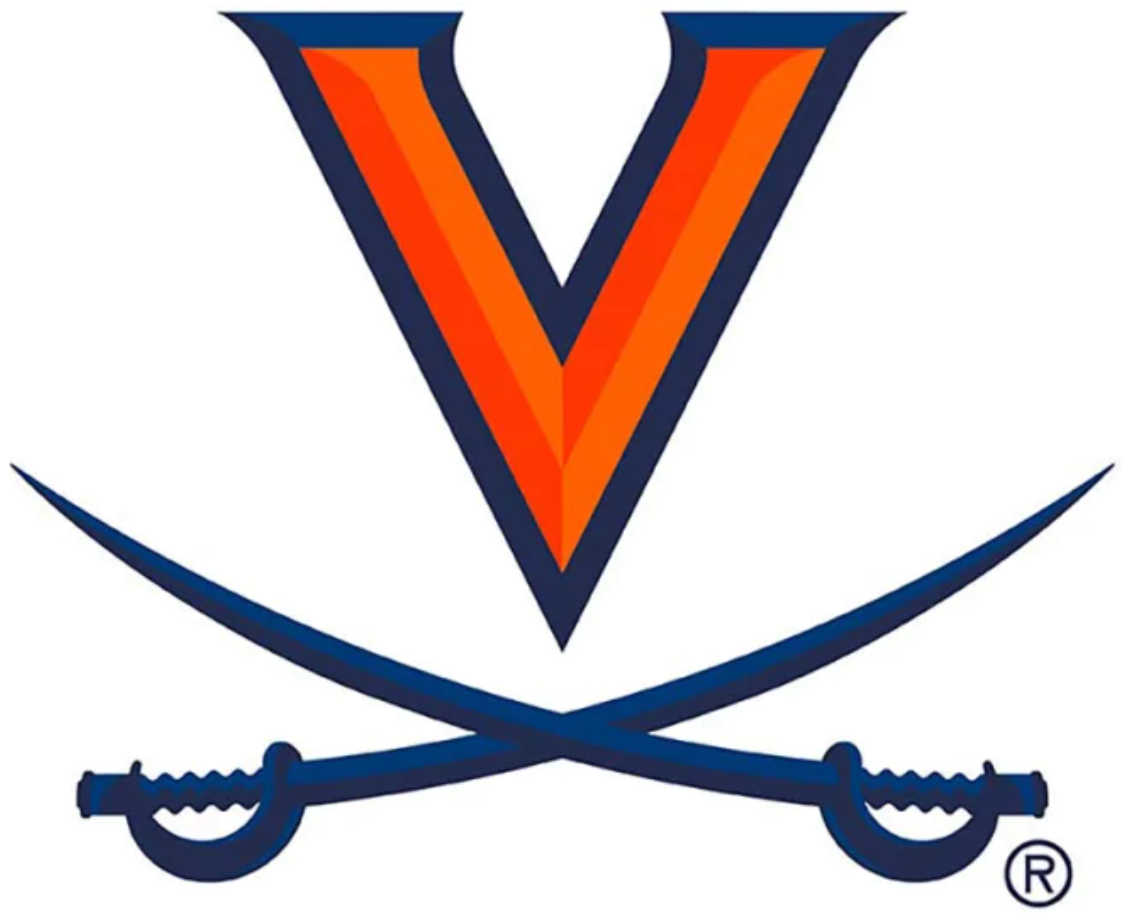}}}

\author[\metamark]{Di Liu}
\author[\metamark\,\uvamark]{Xuweiyi Chen}
\author[\metamark]{Cheng Zhang}
\author[\metamark]{Adam Harley}
\author[\metamark]{Zhao Dong}
\author[\metamark]{Tanner Schmidt}
\author[\metamark]{David Caruso}
\author[\metamark]{Pierre Moulon}
\author[\metamark]{Shugao Ma}
\author[\uvamark]{Zezhou Cheng}
\author[\metamark]{Richard Newcombe}
\author[\metamark]{Jakob Julian Engel}

\affiliation[\metamark]{Reality Labs Research, Meta}
\affiliation[\uvamark]{University of Virginia}

\definecolor{projectlinkcolor}{HTML}{112D4E}
\definecolor{codelinkcolor}{HTML}{3F72AF}
\definecolor{benchmarklinkcolor}{HTML}{846491}
\newcommand{\projectlinks}{{\small\sffamily
  \href{https://facebookresearch.github.io/rlr_scale4d/}{\textcolor{projectlinkcolor}{\faGlobe\enspace Project page}}\quad
  \href{https://github.com/facebookresearch/rlr_scale4d}{\textcolor{codelinkcolor}{\faGithub\enspace Code}}\quad
  \href{https://huggingface.co/facebook/SCALE4D}{\textcolor{benchmarklinkcolor}{\faDatabase\enspace Benchmark}}%
}}
\patchcmd{\mymaketitle}
  {\affiliationlist\par}
  {\affiliationlist\par\vskip 0.2cm\projectlinks\par}
  {}{\PackageError{scale4d}{Could not insert project links below affiliations}{Check the fairmeta title layout.}}

\abstract{
Metric 3D geometry and motion are essential for embodied reasoning.
Yet existing feedforward 4D models underuse
synchronized views and calibrated cameras available to embodied systems. We introduce \themodel, a unified feedforward model that predicts metric 3D geometry and trajectories from single- or multiview videos, with optional camera conditioning. 
A shared point-query representation integrates spatiotemporal attention and cross-view fusion, while metric camera positional encoding incorporates calibrated geometry without compromising physical scale. 
Geometry-preserving scale augmentation encourages the model to use camera baselines for metric prediction. 
Beyond supervised learning, \themodel's flexible camera conditioning enables asymmetric self-training, where stronger camera-conditioned predictions supervise RGB-only inference on unlabeled videos.
Across egocentric and robotic
benchmarks, \themodel achieves strong metric tracking and depth without ground-truth
scale alignment. We further demonstrate zero-shot generalization to larger
numbers of views, with performance continuing to improve
beyond the maximum of three views seen during training.
}

\date{\today}

\makeatletter
\providecommand\@LN[2]{}
\providecommand\@LN@col[1]{}
\renewcommand\paragraph{\@startsection{paragraph}{4}{\z@}%
  {1.25ex \@plus 1ex \@minus .2ex}%
  {-.5em}%
  {\normalfont\normalsize\bfseries}} %
\makeatother

\begin{document}

\maketitle
\setcounter{footnote}{1}
\newcommand{\teasercaption}{\textbf{\themodel} predicts dense metric 3D trajectories across
embodied perception scenarios, supporting varying
view counts and optional camera conditioning.
Cones~(\raisebox{-0.1em}{\includegraphics[height=0.8em]{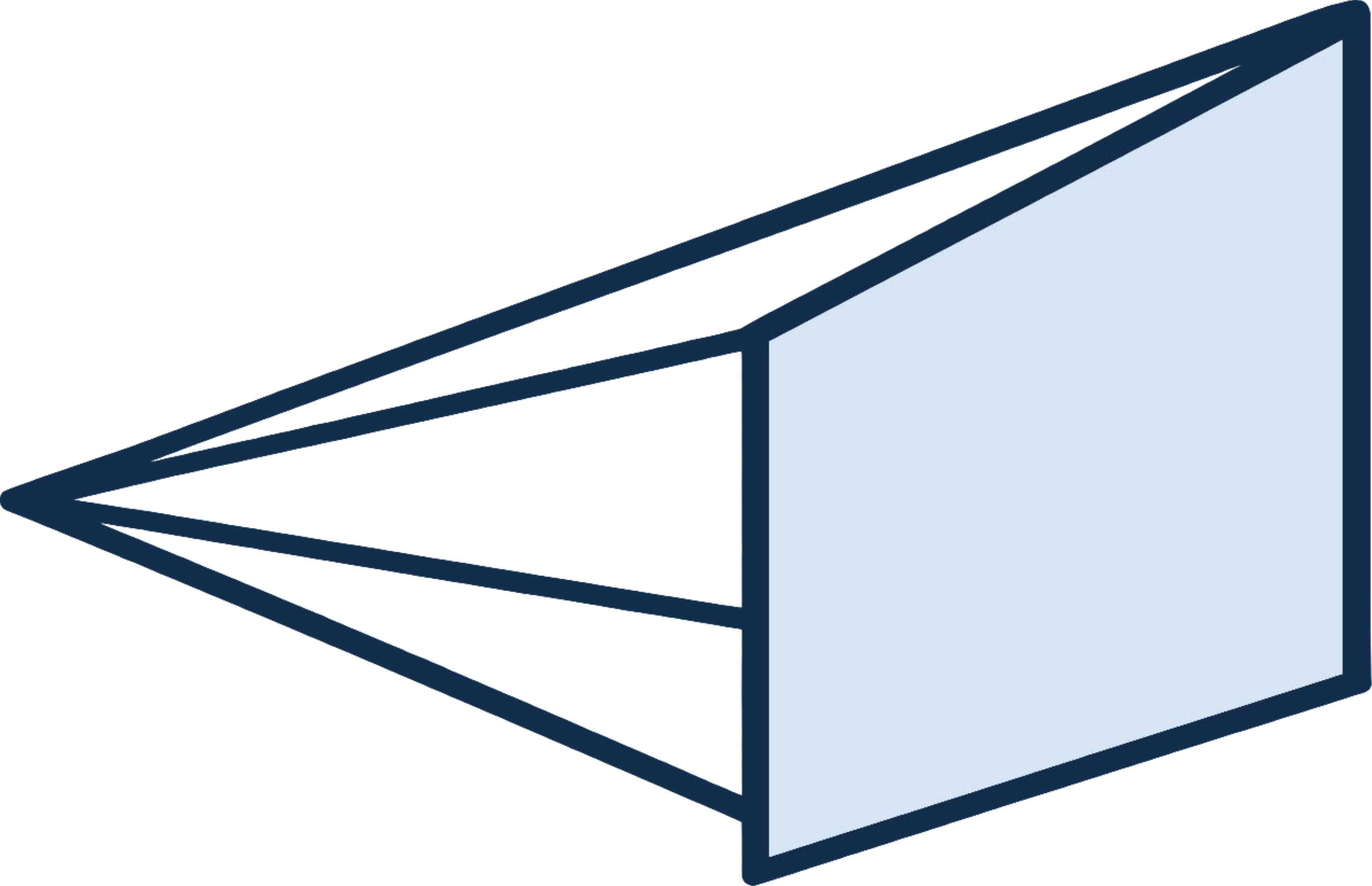}})
show supported view counts;
slashed circles~(\raisebox{-0.1em}{\includegraphics[height=0.8em]{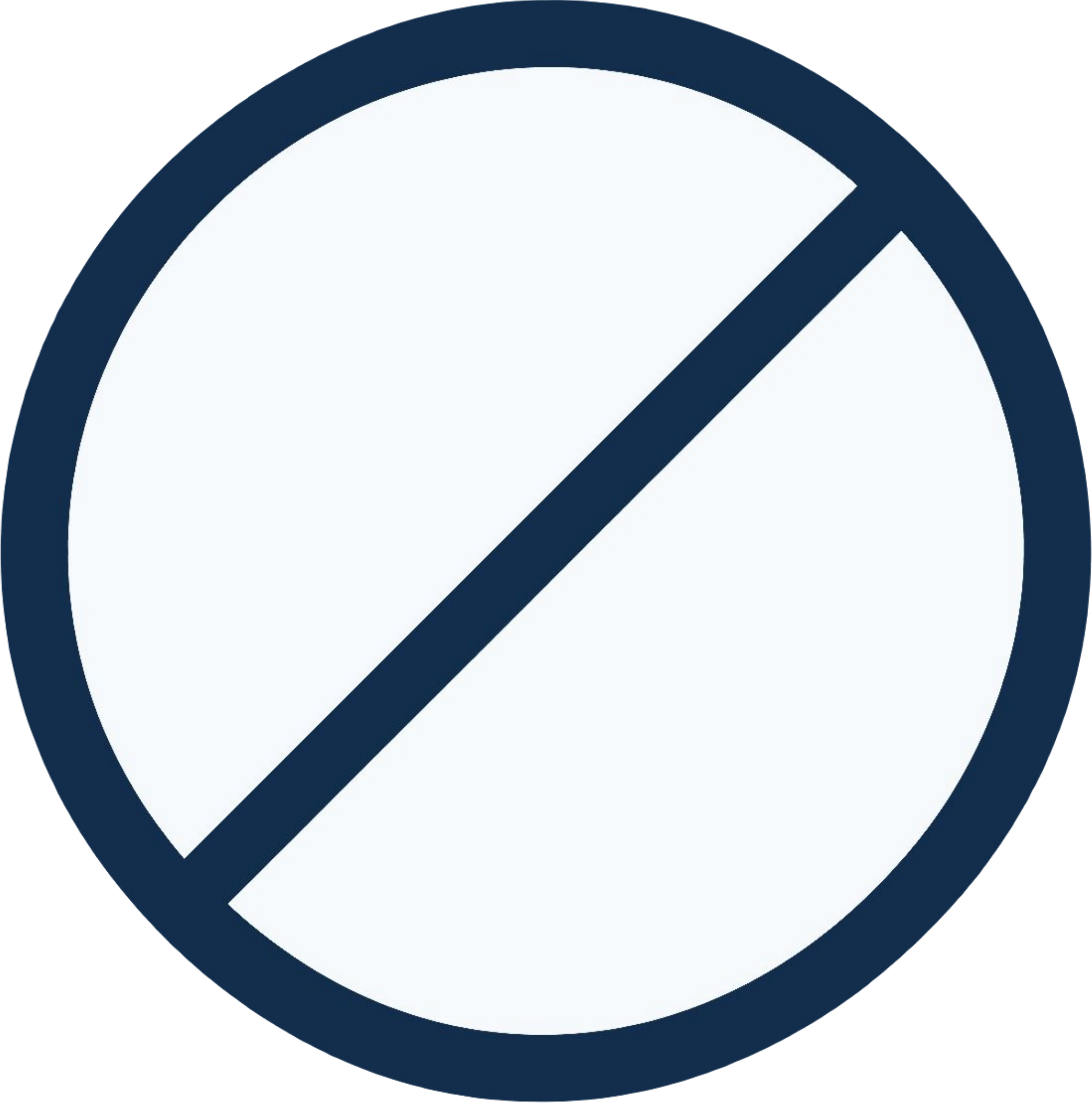}})
mark disabled camera conditioning. Numbers indicate predicted 3D distances (cm) between marked points.
Filmstrips present frames from the query view.
}
\ifdefined\teaserfirstpage
\begingroup
    \centering
    \vspace{-20pt}
    \captionsetup{hypcap=false,skip=2pt}
    \noindent\includegraphics[width=\textwidth,trim=0 1.5pt 0 1pt,clip]{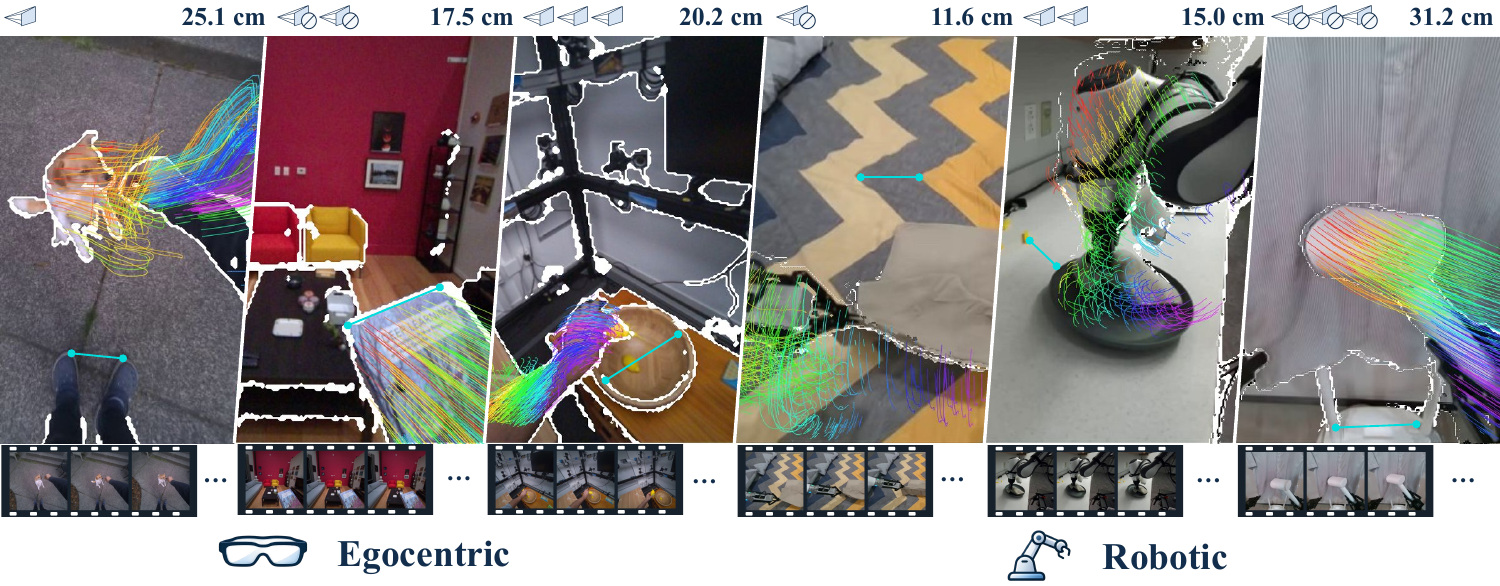}\par
    \captionof{figure}{\teasercaption}
    \label{fig:teaser}
    \par
    \vspace*{10pt}
\endgroup
\else
\begin{figure*}[t]
    \centering
    \captionsetup{skip=3pt}
    \includegraphics[width=\textwidth,height=210pt,keepaspectratio]{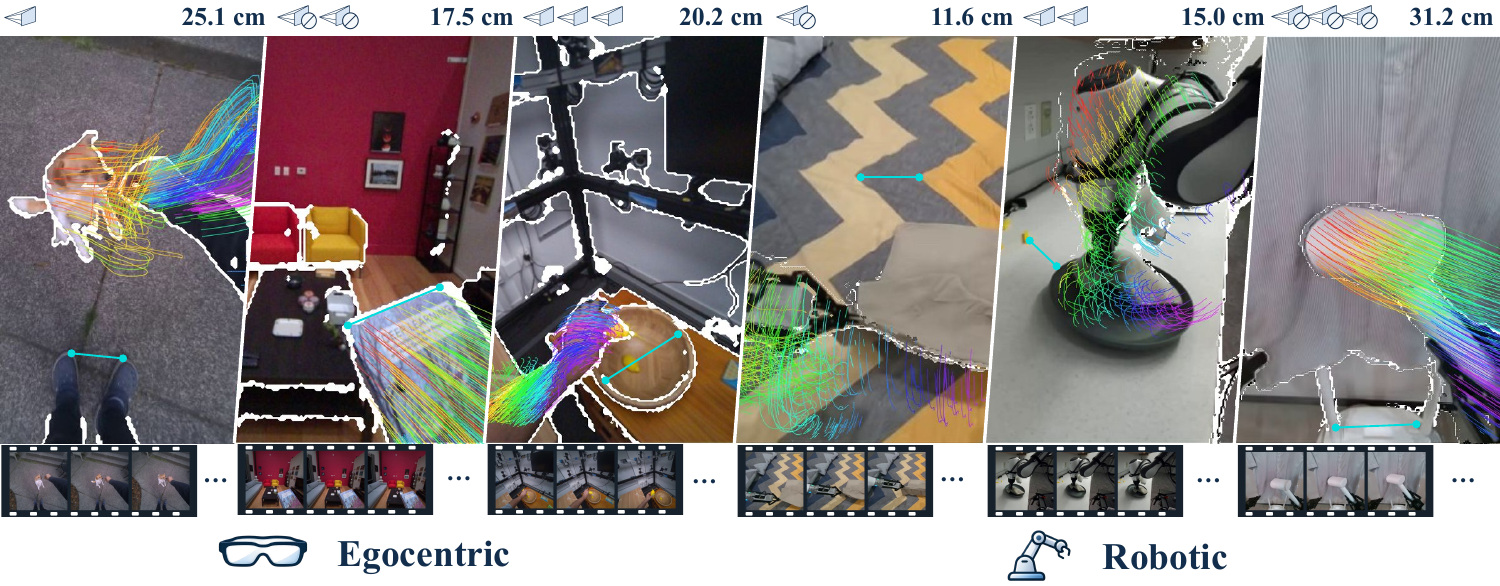}
    \caption{\teasercaption}
    \label{fig:teaser}
\end{figure*}
\fi

\section{Introduction}
\label{sec:intro}

Embodied systems, from wearable devices to robots, should
understand not only what happens around them, but also
where objects are and how they move in metric space~\cite{zhong2026beyond}.
Metric 3D geometry and motion are essential for embodied reasoning, enabling physically meaningful estimates of distances, velocities, and accelerations needed to understand and interact with the environment.
Recent feedforward 4D models enable direct prediction of
scene geometry and trajectories from
video~\cite{feng2025st4rtrack,sucar25dynamic,
xiao2025spatialtrackerv2,zhang2026d4rt,liu2025traceanything},
yet recovering accurate metric geometry remains challenging.
Unlike general video, embodied systems often capture
synchronized observations from multiple cameras, together
with calibrated camera poses~\cite{engel2023aria,banerjee2024hot3d,khazatsky2024droid}.
These measurements provide complementary geometric evidence
and physical scale references that could improve 4D perception~\cite{jang25pow3r:,keetha2026mapanything}.
However, existing approaches have limited flexibility in
leveraging such measurements across different sensing
configurations.

We introduce \themodel, a unified framework for metric 4D
perception that adapts to the geometric information available
at inference (Fig.~\ref{fig:teaser}).
Our model supports both single-view and multiview videos,
with or without calibrated camera information, using a
shared representation and a single set of weights.
These configurations provide different levels of geometric
evidence: additional views offer complementary observations
that help resolve spatial ambiguity, while calibrated
camera poses provide a physical reference for recovering
metric scale.
Rather than treating these settings as separate tasks,
\themodel{} integrates visual and geometric evidence within
a unified feedforward architecture, enabling metric geometry
and motion prediction across diverse embodied scenarios.

A central challenge is reliably incorporating calibrated
camera geometry into metric 4D prediction. Directly using
camera translations in physical units can destabilize
geometric attention during training, while normalizing them risks losing metric scale information needed for accurate prediction.
We address this challenge through metric camera positional
encoding (MCPE), which regulates camera translation magnitudes
while explicitly retaining metric scale cues.
We further introduce geometry-preserving scale augmentation
to encourage the model to use calibrated camera baselines
rather than relying solely on appearance-based scale priors.

Beyond its strong metric predictions, \themodel's flexible input interface also provides a natural foundation for self-training. In particular, calibrated cameras enable more accurate metric predictions that can supervise RGB-only learning on unlabeled videos. We explore this capability through camera-asymmetric self-training, where a camera-conditioned teacher supervises an RGB-only student on unlabeled videos. We show that this asymmetry improves camera-free inference over symmetric self-training, illustrating how geometric measurements can benefit learning even when they are not available at inference.

Our contributions are: (1) a unified feedforward framework for metric 4D perception that supports flexible view counts and camera configurations, generalizing zero-shot from at most three training views to ten; (2) MCPE and geometry-preserving scale augmentation for stable and accurate use of metric camera information; and 
(3) a demonstration of \themodel's potential for self-training through camera-asymmetric supervision, improving RGB-only metric prediction using unlabeled calibrated videos.
\section{Related Work}
\label{sec:related-work}

\noindent\textbf{Metric Space Modeling.}
Recovering geometry in physical units remains challenging due to the inherent scale ambiguity of monocular observations.
Monocular approaches such as
Metric3D~\cite{yin2023metric3d},
UniDepth~\cite{piccinelli2024unidepth,piccinelli2025unidepthv2},
and MoGe~\cite{wang2025moge2,kong2026moge3} learn metric geometry from RGB images,
while Depth Anything 3~\cite{depthanything3} extends geometric prediction to
flexible view configurations.
Alternatively, calibrated stereo provides a direct
metric reference for depth
estimation~\cite{wen2025foundationstereo},
and methods such as Pow3R~\cite{jang25pow3r:}
and MapAnything~\cite{keetha2026mapanything}
incorporate additional geometric measurements.
Extending metric prediction to dynamic scenes is more
challenging, as both scene geometry and motion must be
recovered in a consistent metric coordinate system.
Any4D~\cite{karhade2025any4d} addresses this problem
through a factorized representation, predicting
scale-normalized geometry and motion together with
a separate global metric scale factor.
In contrast, \themodel{} directly predicts metric
3D positions and trajectories through a unified
point--time query representation, incorporating calibrated
camera geometry into attention when available.\looseness=-1

\noindent\textbf{Unified 4D Reconstruction and Tracking.}
Feedforward reconstruction provides an alternative to
classical structure-from-motion and multiview
stereo~\cite{schoenberger2016sfm,schoenberger2016mvs}.
DUSt3R~\cite{wang24dust3r:} and
MASt3R~\cite{leroy2024mast3r} regress point maps
from image pairs, while VGGT~\cite{wang25vggt}
and related methods~\cite{yang25fast3r:,wang25p3}
extend reconstruction to multiple images.
Recent approaches recover dynamic geometry from
video~\cite{zhang24monst3r:,lu2024align3r,li2024megasam,chen2025easi3r,wang2025c4d,wang2025cut3r},
integrate geometric observations across
time~\cite{sucar25dynamic,sucar2026vdpm,han2025d2ust3r},
or optimize scene-specific dynamic
representations~\cite{wang2024shapeofmotion,lei2024mosca}.
Point trackers such as PIPs~\cite{harley2022pips}, TAPIR~\cite{doersch2023tapir},
CoTracker~\cite{karaev24cotracker}, and
AllTracker~\cite{harley2025alltracker}
predict image-space trajectories, while
SpatialTracker~\cite{xiao2024spatialtracker,xiao2025spatialtrackerv2},
DELTA~\cite{ngo2024delta}, and
TAPIP3D~\cite{tapip3d}
recover 3D trajectories using depth information.
Other methods directly predict 3D
motion~\cite{feng2025st4rtrack,lu2026track4world,luo20264rc,jeon2026point4d},
while multiview tracking
approaches~\cite{rajic2025mvtracker,koo2026mvtap,jiang2026omnix,fradlin2026arrow}
exploit synchronized observations.
More recently, D4RT~\cite{zhang2026d4rt},
L4P~\cite{badki2025l4p}, and
TraceAnything~\cite{liu2025traceanything}
unify geometry and motion prediction within
feedforward 4D frameworks.
\themodel{} extends unified 4D perception to metric
space with flexible view counts and optional
camera conditioning, without requiring an external
depth estimator.

\noindent\textbf{Camera Conditioning and Geometric Attention.}
Per-pixel Pl\"ucker embeddings encode viewing
rays~\cite{sitzmann2021lfn,he2024cameractrl},
although their absolute coordinates depend on the
chosen world frame.
In geometric attention, GTA~\cite{miyato2024gta}
applies group actions to attention features,
PRoPE~\cite{li2025prope}
introduces projective relative positional encoding,
and UCPE~\cite{zhang2026ucpe} uses local ray frames
to represent relative camera geometry for video generation.
While these methods encode camera relationships,
their application to metric 4D prediction requires
handling camera translations in physical units
without losing absolute scale information.
\themodel{} addresses this challenge through
scale-conditioned geometric attention, enabling
stable metric camera conditioning alongside
RGB-only inference.\looseness=-1

\paragraph{Self-supervised geometry learning and distillation.}
Learning from unlabeled visual data has been explored through geometric consistency and self-training. Self-supervised depth methods exploit photometric and geometric constraints to learn scene structure without dense annotations~\cite{godard2019digging,bian2019unsupervised}. More recent approaches leverage pseudo-labels for monocular depth estimation~\cite{yang2024depth} and point tracking~\cite{doersch2024bootstap}. Selfi~\cite{deng2026selfi} improves 3D reconstruction through geometric feature alignment, while SelfEvo~\cite{huang2026self} introduces online self-distillation with spatiotemporal context asymmetry for 4D perception. \themodel instead exploits geometric observation asymmetry, where camera conditioning enable a teacher to supervise metric trajectory prediction from reduced inputs. Supervised replay preserves a metric reference as self-labeling expands to additional calibrated videos.

\section{Metric 4D Modeling}
\label{sec:method}

\begin{figure*}[!t]
    \centering
    \includegraphics[width=0.98\textwidth]{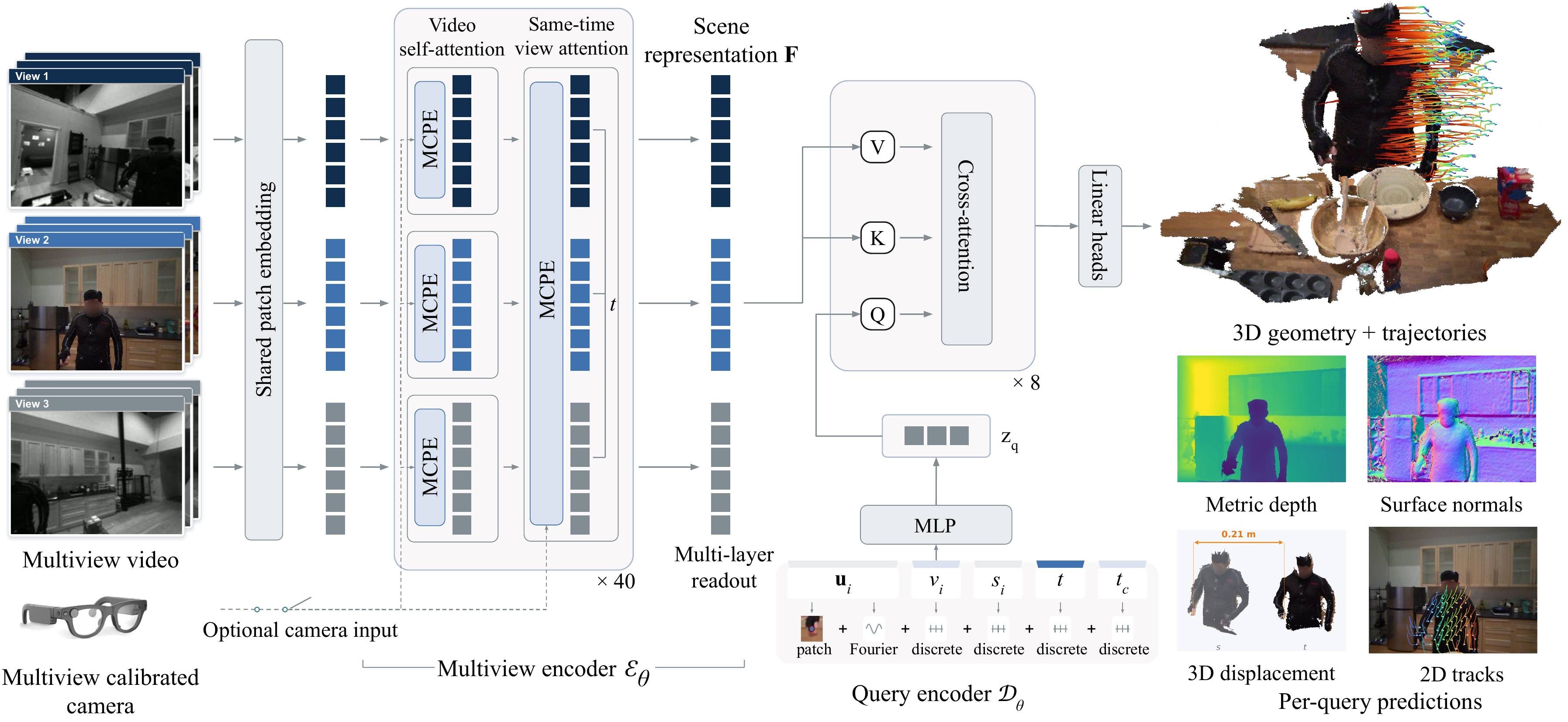}
    
    \caption{\textbf{\themodel architecture.}
    Video self-attention within each view and same-time attention across views
    encode synchronized videos, with optional MCPE camera conditioning. A
    multi-layer readout forms the scene representation $\mathbf F$. Point--time
    queries decode metric geometry, trajectories, and auxiliary outputs without
    re-encoding the video.}

    \label{fig:method_overview}
    \label{fig:method_encoder_decoder}
\end{figure*}

\themodel extends D4RT's feedforward encoder--decoder architecture~\cite{zhang2026d4rt} to predict metric scene geometry and motion from varying numbers of synchronized views, with optional camera conditioning (Fig.~\ref{fig:method_overview}). Our design exploits calibrated camera geometry when available while supporting prediction from reduced inputs. A central challenge is preserving metric scale: jointly scaling scene geometry and camera translations leaves image observations unchanged, while calibrated camera baselines provide an explicit metric reference that can resolve this ambiguity. We therefore develop a multiview architecture with geometry-aware attention, together with metric learning strategies that preserve and exploit physical scale.\looseness=-1

\subsection{Multiview Architecture}
\label{sec:method_multiview_encoder}

\noindent\textbf{Multiview Spatiotemporal Encoder.}
Given $V$ synchronized videos with $T$ frames
$\mathcal I=\{I_{v,t}\}_{v=1,t=1}^{V,T}$,
a shared embedding layer forms spatial tokens spanning
two consecutive frames.
With weights shared across views, the encoder
$\mathcal E_\theta$ alternates spatiotemporal
\emph{video self-attention} within each view and
\emph{same-time view attention} across views.
With one view, the latter reduces to spatial attention.
The encoder produces multiview features
$\{\mathbf F_v\}_{v=1}^{V}$.

\noindent\textbf{Camera Conditioning.}
To incorporate camera geometry, we introduce a parallel
geometry-aware attention branch following
UCPE~\cite{zhang2026ucpe}, termed metric camera positional
encoding (MCPE).
Given calibration
$\mathcal C=\{\mathbf K_{v,t},\mathbf T^w_{c,v,t}\}$
with camera intrinsics and metric camera-to-world poses,
MCPE associates video tokens with viewing rays and
operates alongside RGB attention in both encoder stages.
Omitting MLPs, each stage updates token features
$\mathbf z=\{\mathbf z_a\}$, where
$\mathbf z_a\in\mathbb R^D$ denotes the $a$-th token
feature, as
\begin{equation}
\mathbf z'_a
=\mathbf z_a+\mathcal A_{\mathrm{RGB}}(\mathbf z)_a
+b_{\mathrm{cam}}\operatorname{MCPE}(\mathbf z;\mathcal C)_a,
\label{eq:camera_residual}
\end{equation}
where $\mathcal A_{\mathrm{RGB}}$ denotes standard
attention over video token features, and
$b_{\mathrm{cam}}\in\{0,1\}$ enables or disables
camera conditioning.
We jointly train both settings, allowing the same model
weights to operate with or without camera information.
Sec.~\ref{sec:method_ucpe} details MCPE.

\noindent\textbf{Metric 4D Query Decoder.}
Following D4RT~\cite{zhang2026d4rt}, we represent
4D geometry through point--time queries.
A query
$q_{i,t}=(\mathbf u_i,v_i,s_i,t,t_c)$
identifies point $i$ by its source pixel
$\mathbf u_i\in\mathbb R^2$, source view $v_i$,
and source time $s_i$, and requests its 3D position
at target time $t$ in the reference camera frame
$(v_i,t_c)$.
To construct shared decoder memory $\mathbf F$,
we normalize and concatenate encoder features
across layers for each view, project them into tokens,
and aggregate tokens across views.
A query decoder $\mathcal D_\theta$ predicts
$
\hat{\mathbf x}_{i,t}^{v_i,t_c}
=\mathcal D_\theta(q_{i,t};\mathbf F)
\in\mathbb R^3.
$
The decoder uses cross-attention to retrieve scene
features, together with a local RGB patch encoding
the source point's appearance.
Fixing $(\mathbf u_i,v_i,s_i,t_c)$ and varying $t$
yields a metric 3D trajectory.
Querying all pixels at $t=s_i=t_c$ produces a depth
map, while fixing $t=s_i$ and a shared reference
frame yields a 3D point cloud.
Auxiliary heads predict image positions, visibility,
3D motion, normals, and confidence.
Each clip is encoded once, and all queries reuse
its features.

\subsection{Metric Camera Positional Encoding}
\label{sec:method_ucpe}

We build on UCPE~\cite{zhang2026ucpe}, which incorporates
camera geometry into attention through relative ray
transformations. For each image token $a$, the camera
origin, viewing direction, and camera's downward axis
define a local ray frame with ray-to-world transformation
$\mathbf T_a^{\mathrm{wr}}$, where $w$ and $r$ denote
the world and ray coordinate frames, respectively.
The relative ray transformation between tokens $a$ and $b$
is $\mathbf G_{ab}=(\mathbf T_a^{\mathrm{wr}})^{-1}
\mathbf T_b^{\mathrm{wr}}$.
UCPE constructs token-dependent operators
$\mathbf D_a=\operatorname{blkdiag}
(\mathbf D_a^{\mathrm{ray}},\mathbf D_a^{\mathrm{RoPE}})$,
where $\mathbf D_a^{\mathrm{ray}}$ applies the
world-to-ray transformation
$(\mathbf T_a^{\mathrm{wr}})^{-1}$
to four-channel geometric feature groups, while
$\mathbf D_a^{\mathrm{RoPE}}$ encodes spatial positions.
The geometric components satisfy\looseness=-1
\begin{equation}
\mathbf D_a^{\mathrm{ray}}
(\mathbf D_b^{\mathrm{ray}})^{-1}
=\mathbf I\otimes\mathbf G_{ab},
\label{eq:relative_ray_operator}
\end{equation}
where $\mathbf I$ is an identity matrix over
geometric feature groups.
The resulting attention is
\begin{equation}
\begin{aligned}
A_{ab}&=\operatorname{softmax}_{b}\!\left(
\frac{\mathbf q_a^\top
\mathbf D_a\mathbf D_b^{-1}\mathbf k_b}{\sqrt d}
\right),\\
\mathbf o_a&=\mathbf D_a
\sum_b A_{ab}\mathbf D_b^{-1}\mathbf v_b,
\end{aligned}
\label{eq:ucpe_attention_a}
\end{equation}
where $A_{ab}$ is the attention weight from
query token $a$ to key token $b$,
$\mathbf q_a$, $\mathbf k_b$, and $\mathbf v_b$
are the query, key, and value of head
dimension $d$, and $\mathbf o_a$ is the attention
output for token $a$.

\begin{figure}[t]
    \centering
    \includegraphics[width=\columnwidth]{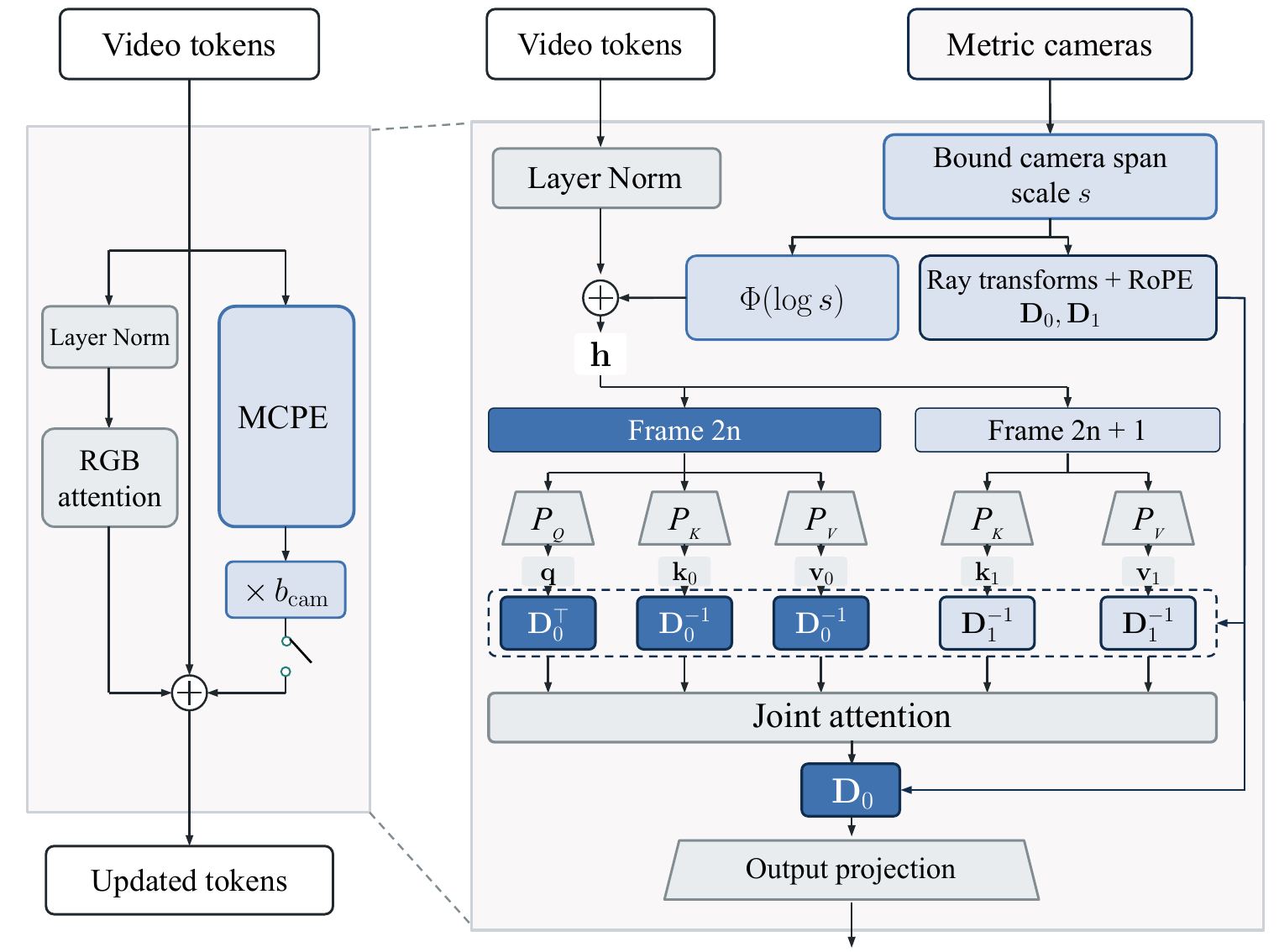}
    \caption{\textbf{Overview of MCPE integration.}
    MCPE runs beside RGB attention as a residual gated by $b_{\mathrm{cam}}$.
    Each two-frame token gives one attention query $\mathbf q$ and a key--value entry
    $(\mathbf k_r,\mathbf v_r)$ per frame $r$, with shared projections and frame
    ray transforms $\mathbf D_r$, and both entries share one softmax.}
    \label{fig:scale_conditioned_camera}
\end{figure}

Applying this formulation to metric 4D prediction
presents two challenges. First, camera translations in
physical units can produce large attention logits,
making geometric conditioning numerically unstable.
Second, each video token spans two consecutive frames
with potentially different camera poses, whereas
UCPE associates each token with a single ray frame.
We address these challenges through scale-conditioned
and dual-frame ray attention
(Fig.~\ref{fig:scale_conditioned_camera}).

\noindent\textbf{Scale-conditioned Attention.}
As shown in Eq.~\eqref{eq:relative_ray_operator},
each four-channel geometric feature group is
transformed by the relative ray transformation
$\mathbf G_{ab} = (\mathbf R_{ab}, \mathbf t_{ab})$,
where $\mathbf R_{ab}$ and $\mathbf t_{ab}$
denote relative rotation and translation.
For a query--key pair
$\mathbf q=(\mathbf q_{xyz},q_h)$ and
$\mathbf k=(\mathbf k_{xyz},k_h)$, the contribution
to the attention logit in Eq.~\ref{eq:ucpe_attention_a} expands as
\begin{equation*}
\begin{aligned}
\mathbf q^\top\mathbf G_{ab}\mathbf k
&=\mathbf q_{xyz}^\top\mathbf R_{ab}\mathbf k_{xyz}\\
&\quad+k_h\mathbf q_{xyz}^\top\mathbf t_{ab}
+q_hk_h.
\end{aligned}
\label{eq:camera_translation_logit}
\end{equation*}
The translation-dependent term $k_h\mathbf q_{xyz}^\top\mathbf t_{ab}$ scales with camera
separation, potentially producing large attention
logits when using metric camera poses.

To regulate this dependence while retaining metric
scale information, we normalize camera translations
and separately encode their original scale.
Let
$\Delta=\max_{j,k}\|\mathbf c_j-\mathbf c_k\|_2$
denote the maximum camera separation across all
views and times. We define
\begin{equation}
s=f(\Delta/C),\qquad
\bar{\mathbf c}_j=
(\mathbf c_j-\mathbf c_{\mathrm{ref}})/s,
\end{equation}
where $f(x)=\max(1,x)$, $C=20$\,m, and
$\mathbf c_{\mathrm{ref}}$ is the first camera center.
Camera spans up to $C$ thus keep their metric
translations ($s=1$), while larger spans are
scaled down to $C$.
We construct the ray transformations from the
normalized camera centers while keeping rotations
and intrinsics unchanged.

To retain the original metric scale, we condition the MCPE attention features on the normalization factor through a learned bias-free projection $\Phi:\mathbb{R}\rightarrow\mathbb{R}^D$, yielding $\mathbf{h}=\operatorname{LN}_{\mathrm{cam}}(\mathbf{z})+\Phi(\log s)$.
The scale embedding is shared across tokens,
and $\mathbf h$ is projected into the queries,
keys, and values of the MCPE attention branch.
Thus, normalized ray geometry regulates
translation-dependent attention logits, while
scale-conditioned features retain the original
metric scale of the camera layout.

\noindent\textbf{Dual-frame Ray Attention.}
Each video token jointly represents two consecutive frames, which may have different camera poses. Associating the token with only one ray frame would discard the geometric information of the other frame. We therefore construct two key--value entries from the same token representation, each associated with its corresponding frame's ray geometry. The entries share projection weights, while queries use the first frame's ray coordinate system.\looseness=-1

We extend
Eq.~\eqref{eq:ucpe_attention_a} to two temporal
ray frames per token. For token $a$, let
$\mathbf D_{a,0}$ and $\mathbf D_{a,1}$
denote the geometric operators of its two frames,
constructed from normalized camera centers.
Both key--value entries share the same
scale-conditioned feature $\mathbf h_a$ and
projection weights, with frame-specific Lat-Up
biases added when gravity information is available.
Queries use the first frame's ray representation,
while attention is jointly normalized over
tokens and frames:
\[
\begin{aligned}
A_{abr}&=\operatorname{softmax}_{b,r}\!\left(
\frac{\mathbf q_a^\top \mathbf D_{a,0}
\mathbf D_{b,r}^{-1}\mathbf k_{b,r}}{\sqrt d}
\right),\\
\mathbf o_a&=\mathbf D_{a,0}\sum_b\sum_{r=0}^{1}
A_{abr}\mathbf D_{b,r}^{-1}\mathbf v_{b,r},
\end{aligned}
\]
where $r\in\{0,1\}$ indexes the two frames.
Restricting attention to $r=0$ recovers the
single-frame formulation of Eq.~\eqref{eq:ucpe_attention_a}.

\subsection{Learning Metric Scale and Training}
\label{sec:method_training}
\label{sec:method_scale_augmentation}

\noindent\textbf{Scale Augmentation.}
Metric supervision alone does not ensure that the model
uses calibrated camera geometry rather than appearance-based
scale priors. To encourage geometric scale reasoning,
we jointly rescale camera translations and scene geometry
while keeping RGB observations unchanged.
For a sampled factor $\alpha>0$, camera centers $\mathbf c_j$
and world-space target points $\mathbf X$ are scaled about
the reference camera center $\mathbf o=\mathbf c_{\mathrm{ref}}$:
$\mathbf c_j^\alpha=\mathbf o+\alpha(\mathbf c_j-\mathbf o)$
and $\mathbf X^\alpha=\mathbf o+\alpha(\mathbf X-\mathbf o)$.
Camera rotations and intrinsics remain unchanged, preserving
image projections while scaling metric 3D positions and
displacements by $\alpha$.
Identical RGB observations thus correspond to different metric
targets, forcing the model to rely on calibrated camera
separations to infer scale.
We apply scale augmentation only to camera-conditioned examples
with sufficient camera displacement, before camera-span normalization.

\noindent\textbf{Training Objectives.}
We replace D4RT's depth-normalized position objective with
$\mathcal L_{\mathrm{3D}}^{\mathrm{metric}}$, a confidence-weighted
$L_1$ loss on metric 3D positions.
Unlike depth-normalized objectives, it directly penalizes
errors in meters.
The total objective combines this loss with auxiliary
supervision for 2D positions, visibility, motion, normals,
and confidence, as well as multiview reprojection:\looseness=-1
\begin{equation}
\begin{aligned}
\mathcal L={}&\mathcal L_{\mathrm{3D}}^{\mathrm{metric}}
+\lambda_{\mathrm{2D}}\mathcal L_{\mathrm{2D}}
+\lambda_{\mathrm{vis}}\mathcal L_{\mathrm{vis}}\\
&+\lambda_{\mathrm{motion}}\mathcal L_{\mathrm{motion}}
+\lambda_{\mathrm{normal}}\mathcal L_{\mathrm{normal}}\\
&+\lambda_{\mathrm{conf}}\mathcal L_{\mathrm{conf}}
+\lambda_{\mathrm{reproj}}\mathcal L_{\mathrm{reproj}}.
\end{aligned}
\label{eq:loss}
\end{equation}
Sec.~\ref{sec:final_recipe} lists the loss weights.
To enforce geometric consistency across calibrated views, we express predicted 3D positions in world coordinates $\hat{\mathbf X}_{i,t}^{w}$, transform them into the target camera coordinate system, and project them into view $v$ at time $t$:\looseness=-1
\begin{equation}
\ell_{\mathrm{reproj}}(i,t,v)=
\left\|
\Pi_{v,t}\!\left(
\mathbf R_{v,t}^{cw}\hat{\mathbf X}_{i,t}^{w}
+\mathbf t_{v,t}^{cw}
\right)
-\mathbf u^*_{i,t,v}
\right\|_1,
\label{eq:reprojection_error}
\end{equation}
where $\mathbf R_{v,t}^{cw}$ and $\mathbf t_{v,t}^{cw}$ define the world-to-camera transformation, $\Pi_{v,t}$ projects camera-space points into image coordinates using the camera intrinsics, and $\mathbf u^*_{i,t,v}$ denotes the observed correspondence. Both projected and observed image coordinates are normalized by image size.
We average valid reprojection errors separately over the source and other calibrated views, summing the two contributions.
Only visible, in-frame correspondences with positive predicted depth are considered, and cross-view errors are clipped to reduce the influence of outliers.
Importantly, supervision views need not be encoder inputs, allowing additional geometric observations to constrain metric predictions during training.

\subsection{Camera-Asymmetric Self-Training}
\label{sec:self_distillation}
\label{sec:data_flywheel}

\begin{figure}[!t]
    \centering
    \includegraphics[width=0.98\linewidth]{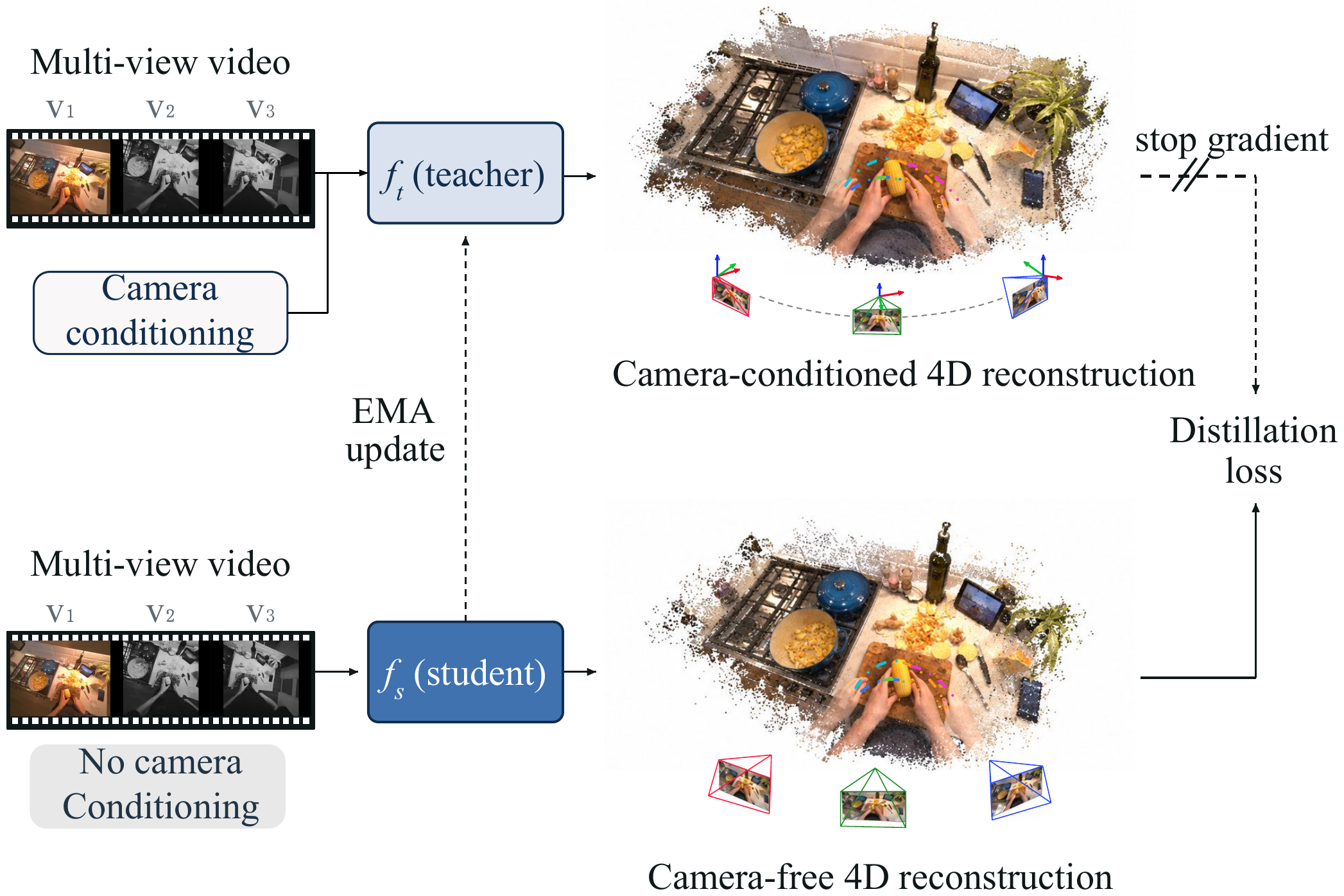}
    \caption{\textbf{Camera-asymmetry self-training.}
    A camera-conditioned teacher supervises a student without camera conditioning using the same input views.
    The teacher generates metric trajectory targets for the student, with its weights updated through an exponential moving average (EMA) of the student.}
    \label{fig:data_flywheel}
\end{figure}

The strong metric predictions and flexible camera conditioning of \themodel make it a natural foundation for self-training. In particular, camera-conditioned
predictions can provide geometric supervision for RGB-only inference on unlabeled calibrated videos.
We exploit this capability through an asymmetric teacher--student setup, where the teacher receives
camera information while the student operates on the same RGB observations without cameras(Fig.~\ref{fig:data_flywheel}).

The teacher and student share the same architecture,
pretrained initialization, and RGB observations, but only
the teacher receives calibrated camera information.
This allows the teacher to provide calibration-informed
metric supervision for RGB-only prediction.
For shared point--time queries, the student learns from
detached teacher predictions of 3D positions, motion,
and auxiliary geometric quantities in the same metric
coordinate frame, without scale or rigid alignment.
Teacher confidence weights valid pseudo-targets in
the distillation objective $\mathcal L_{\mathrm{distill}}$.

To stabilize continued learning, we interleave distillation
on unlabeled videos with supervised replay from labeled
data, sampling the two streams with probabilities 0.1
and 0.9, respectively.
We update the teacher as an exponential moving average
(EMA) of the student:
$\theta_{\mathrm T}\leftarrow
\tau\theta_{\mathrm T}+(1-\tau)\theta_{\mathrm S}$,
where $\tau$ is the EMA decay.
This enables continued learning from calibrated videos
without requiring additional ground-truth 3D trajectories.
Sec.~\ref{sec:exp_flywheel} compares it with symmetric self-training
on ADT and HOT3D, evaluating the student with cameras both off and on.

\section{Experiments}
\label{sec:experiments}

\subsection{Training Data}
\label{sec:data}
\label{sec:data_mixture}
\label{sec:data_labeling}
\label{app:data}
\label{app:data_sources}
\label{app:data_labeling}

Our training mixture contains public and internal
datasets including Dynamic Replica~\cite{karaev23dynamicstereo},
PointOdyssey~\cite{zheng2023pointodyssey},
FallingThings~\cite{tremblay2018fallingthings},
Kubric~\cite{greff2022kubric},
TartanAir~\cite{wang2020tartanair},
TartanGround~\cite{patel2025tartanground},
CO3Dv2~\cite{reizenstein2021co3d},
GigaHands~\cite{fu2025gigahands}, and
Waymo Open~\cite{sun2020waymo}.
All clips are standardized into synchronized RGB,
calibrated cameras, metric 3D trajectories,
and validity masks. Appendix~\ref{app:training_sources} lists the
number of samples and frames of each source (Table~\ref{tab:flywheel_data}).

\noindent\textbf{Metric scale filtering.}
Metric supervision is only as reliable as the
scale of its labels.
For sources with uncertain scale, we rescale
each clip using the median ratio between its
depth and frozen DA3 metric depth~\cite{depthanything3}.
We measure these ratios at visible points
and retain only clips with consistent ratios
in log space, filtering clips with unreliable scale estimates
out of the metric supervision.

\noindent\textbf{Stereo trajectory annotation.}
For internal calibrated stereo data, we construct
metric 3D trajectories from stereo depth estimated by
FoundationStereo~\cite{wen2025foundationstereo}
and 2D tracks from AllTracker~\cite{harley2025alltracker}.
Stereo provides metric depth but no temporal
correspondences, while 2D trackers can drift.
Our labeling pipeline therefore retains tracks
that agree across multiple anchors and refines
their depth estimates.

AllTracker runs forward and backward from evenly
spaced anchors.
When a track reaches another anchor, tracking
restarts there and is checked over the overlapping
interval.
We retain the longest consistent segment, merge
compatible segments across anchors, and record
the number of supporting anchors per track.\looseness=-1

Static tracks are fitted to a single world point.
For dynamic tracks, we consider depth candidates
around each tracked pixel to handle foreground-
background ambiguity near depth edges, selecting
a sequence based on image-space proximity and
robust 3D motion smoothness.
Finally, we fill missing depths from neighboring
frames, discard tracks with insufficient reliable
3D observations or abrupt jumps, and combine tracker
visibility with depth-based occlusion checks.

\begin{table*}[t]
\centering
\small
\tabcolsep=0.8em
\resizebox{\linewidth}{!}{
\renewcommand{\arraystretch}{0.88}%
\begin{tabular}{@{}ll ccc ccc ccc ccc@{}}
\toprule
\multirow{2}[2]{*}{\textbf{Method}} & \multirow{2}[2]{*}{\textbf{Depth model}} & \multicolumn{3}{c}{\textbf{ADT}~\cite{pan2023adt}} & \multicolumn{3}{c}{\textbf{HOT3D}~\cite{banerjee2024hot3d}} & \multicolumn{3}{c}{\textbf{DexYCB}~\cite{chao2021dexycb}} & \multicolumn{3}{c}{\textbf{DROID}~\cite{khazatsky2024droid}} \\
\cmidrule(lr){3-5} \cmidrule(lr){6-8} \cmidrule(lr){9-11} \cmidrule(lr){12-14}
 & & AJ~$\uparrow$ & $\text{APD}_{3D}$~$\uparrow$ & OA~$\uparrow$ & AJ~$\uparrow$ & $\text{APD}_{3D}$~$\uparrow$ & OA~$\uparrow$ & AJ~$\uparrow$ & $\text{APD}_{3D}$~$\uparrow$ & OA~$\uparrow$ & AJ~$\uparrow$ & $\text{APD}_{3D}$~$\uparrow$ & OA~$\uparrow$ \\
\midrule
\multirow{4}{*}{MVTracker~\cite{rajic2025mvtracker}}
&  DA3-Metric~\cite{depthanything3} & 27.20 & 39.30 & 89.79 & 25.90 & 44.36 & 75.41 & 62.19 & 79.08 & 84.85 & 18.67 & 45.47 & 57.41 \\
&  DA3-Nested~\cite{depthanything3} & \second 42.13 & \second 52.99 & \second 92.20 & 19.69 & 33.23 & 80.27 & 62.79 & 77.91 & \third 86.78 & 16.24 & 42.24 & 56.99 \\
&  MoGe-3~\cite{kong2026moge3} & 34.84 & 47.09 & \third 90.93 & \second 35.74 & \second 58.00 & 73.31 & 63.94 & \third 82.40 & 83.63 & \second 23.80 & \second 58.91 & \second 58.49 \\
&  UniDepth-v2~\cite{piccinelli2025unidepthv2} & \third 39.10 & \third 50.92 & 89.73 & \third 34.50 & \third 53.24 & 78.01 & 60.36 & 76.60 & 86.03 & 16.91 & 40.96 & 56.75 \\
\midrule
\multirow{4}{*}{OmniX~\cite{jiang2026omnix}}
&  DA3-Metric & 23.27 & 34.34 & 82.50 & 8.41 & 14.76 & \third 81.46 & 53.77 & 71.23 & 82.44 & 13.08 & 32.82 & 49.50 \\
&  DA3-Nested & 33.37 & 46.32 & 82.50 & 6.85 & 12.42 & \third 81.46 & 51.38 & 67.69 & 82.44 & 12.78 & 32.88 & 49.50 \\
&  MoGe-3 & 27.94 & 40.42 & 82.50 & 7.61 & 13.40 & \third 81.46 & 56.04 & 73.83 & 82.44 & 14.89 & 37.59 & 49.50 \\
&  UniDepth-v2 & 29.77 & 41.84 & 82.50 & 6.69 & 12.12 & \third 81.46 & 55.14 & 72.98 & 82.44 & 10.43 & 26.66 & 49.50 \\
\midrule
\multirow{4}{*}{ARROW~\cite{fradlin2026arrow}}
&  DA3-Metric & 26.47 & 36.51 & 88.89 & 9.33 & 15.67 & \second 86.74 & \third 68.92 & 81.39 & \second 89.16 & 19.85 & 41.19 & \third 57.67 \\
&  DA3-Nested & 27.24 & 35.84 & 88.89 & 7.66 & 13.16 & \second 86.74 & 68.70 & 79.99 & \second 89.16 & 18.01 & 38.01 & \third 57.67 \\
&  MoGe-3 & 28.70 & 39.58 & 88.89 & 8.71 & 14.59 & \second 86.74 & \second 71.39 & \first 84.09 & \second 89.16 & \third 22.21 & \third 46.45 & \third 57.67 \\
&  UniDepth-v2 & 27.87 & 36.84 & 88.89 & 7.78 & 13.25 & \second 86.74 & 64.88 & 75.92 & \second 89.16 & 14.62 & 30.70 & \third 57.67 \\
\cmidrule(l){1-14}
\textbf{Ours}  &  --- & \first 54.45 & \first 66.92 & \first 94.94 & \first 52.18 & \first 70.39 & \first 88.15 & \first 73.77 & \second 83.72 & \first 92.92 & \first 65.17 & \first 98.77 & \first 84.12 \\
\bottomrule
\end{tabular}
}
\caption{\textbf{Multi-view metric 3D tracking.}
All methods use three synchronized views; no ground-truth scale alignment is
applied. MVTracker consumes known cameras and the listed metric depth.
OmniX and ARROW recover scale from that depth. Our camera-conditioned model
uses known cameras and requires no external depth estimator.
}
\label{tab:mv_tracking}
\end{table*}

\begin{table*}[t]
\centering
\small
\tabcolsep=0.8em
\resizebox{\linewidth}{!}{%
\renewcommand{\arraystretch}{0.80}%
\begin{tabular}{@{}l ccc ccc ccc ccc@{}}
\toprule
\multirow{2}[2]{*}{\textbf{Method}} & \multicolumn{3}{c}{\textbf{ADT}~\cite{pan2023adt}} & \multicolumn{3}{c}{\textbf{HOT3D}~\cite{banerjee2024hot3d}} & \multicolumn{3}{c}{\textbf{DexYCB}~\cite{chao2021dexycb}} & \multicolumn{3}{c}{\textbf{DROID}~\cite{khazatsky2024droid}} \\
\cmidrule(lr){2-4}\cmidrule(lr){5-7}\cmidrule(lr){8-10}\cmidrule(lr){11-13}
 & AJ~$\uparrow$ & $\mathrm{APD}_{3D}$~$\uparrow$ & OA~$\uparrow$
 & AJ~$\uparrow$ & $\mathrm{APD}_{3D}$~$\uparrow$ & OA~$\uparrow$
 & AJ~$\uparrow$ & $\mathrm{APD}_{3D}$~$\uparrow$ & OA~$\uparrow$
 & AJ~$\uparrow$ & $\mathrm{APD}_{3D}$~$\uparrow$ & OA~$\uparrow$ \\
\midrule
DELTA$^{\dagger}$~\cite{ngo2024delta} & 42.62 & 52.88 & 93.81 & 22.09 & 34.57 & 87.54 & \second 63.81 & 74.51 & 90.08 & 19.51 & 44.37 & 68.78 \\
TAPIP3D (RGBD)$^{\dagger}$~\cite{tapip3d} &
 \third 44.39 & \second 54.69 & 93.18
 & 22.22 & 34.79 & 86.71
 & 62.92 & 74.57 & 89.75
 & 21.44 & 48.62 & 69.87 \\
Point4D$^{\ddagger}$~\cite{jeon2026point4d} & 38.58 & 50.40 & 88.09 & 7.80 & 13.58 & 85.48 & 62.03 & \third 74.93 & 88.66 & 20.92 & 47.37 & 59.08 \\
AllTracker $^{\dagger}$~\cite{harley2025alltracker} &
 \second 44.51 & \third 54.17 & \third 95.02
 & 23.05 & 34.53 & \first 89.42
 & 61.61 & 74.21 & 90.38
 & 23.68 & 47.40 & 67.09 \\
CoTracker3 $^{\dagger}$~\cite{karaev25cotracker3} & 44.13 & 52.99 & \first 95.16 & 22.37 & 32.20 & \third 88.97 & \third 63.45 & 73.11 & \second 91.59 & 22.74 & 34.84 & \second 75.63 \\
Track4World$^{\S}$~\cite{lu2026track4world} &
 33.28 & 43.79 & \second 95.14
 & 9.72 & 16.79 & 88.65
 & 63.17 & 74.77 & \third 91.49
 & \third 23.82 & 49.28 & 70.94 \\
4RC$^{\ddagger}$~\cite{luo20264rc} & 29.37 & 42.73 & 79.04 & 16.79 & 28.10 & 79.70 & 58.62 & \first 76.35 & 82.49 & 11.29 & 28.19 & 40.02 \\
OmniX$^{\ddagger}$~\cite{jiang2026omnix} & 34.89 & 48.44 & 82.91 & 6.02 & 10.63 & 83.28 & 47.34 & 63.60 & 83.12 & 17.88 & 43.66 & 49.20 \\
SpaTracker-v2$^{\ddagger*}$~\cite{xiao2025spatialtrackerv2} &
 43.44 & 52.75 & 94.86
 & \third 24.04 & \third 35.64 & 88.80
 & 60.38 & 72.90 & 89.33
 & 23.74 & \third 49.81 & \third 72.48 \\
Any4D$^{\ddagger}$~\cite{karhade2025any4d} & 27.83 & 38.03 & 89.03 & 22.80 & 34.47 & 83.60 & 53.35 & 68.75 & 84.62 & 9.07 & 24.86 & 45.70 \\
$\pi^4$~\cite{wang2026pi} &
 40.87 & 51.59 & 93.76
 & \second 27.97 & \second 40.03 & 88.39
 & 34.80 & 43.63 & 89.28
 & \second 27.43 & \second 57.25 & 62.21 \\
OpenD4RT$^{\ddagger}$~\cite{zhang2026d4rt,opend4rt} & 22.79 & 32.49 & 90.81 & 6.20 & 13.18 & 77.21 & 57.35 & 69.42 & 90.00 & 19.86 & 42.49 & 67.19 \\
\cmidrule(l){1-13}
\textbf{Ours} & \first 52.21 & \first 64.89 & 94.67 & \first 51.74 & \first 69.87 & \second 89.03 & \first 65.72 & \second 76.10 & \first 92.19 & \first 62.53 & \first 97.04 & \first 83.34 \\
\bottomrule
\end{tabular}%
}
\caption{\textbf{Single-view metric 3D tracking.}
No ground-truth scale alignment. Ours uses known intrinsics and metric
poses without external depth. DA3-Nested~\cite{depthanything3} supplies external depth:
$\dagger$ input/lifting; $\ddagger$ scale recovery.
$\S$: embedded DA3; $*$: test-time optimization.}
\label{tab:main_tracking}

\end{table*}

\subsection{Training Setup}
\label{sec:final_recipe}

We train a single model with one to three input
views and optional camera conditioning.
Each example has 48 frames per view at
$224\times224$ resolution.
Camera conditioning is enabled with probability
0.5, independently of the view count, using a
shared gate across all MCPE branches.
During training, one source view is selected per
example, and the camera reference time $t_c$
equals the target time $t$ with probability 0.4.
At evaluation, we use $t_c=t$.

We apply geometry-preserving scale augmentation
when MCPE is enabled and at least one encoded
camera center is at least 5\,cm from the first.
In these cases, we sample
$\alpha\sim\mathcal U(0.8,1.2)$;
otherwise, $\alpha=1$.
We scale camera translations and metric 3D
supervision while preserving RGB observations
and the 2D image correspondences.
The same RGB frames thus appear with metric
targets that are up to 20\% larger or smaller.

We train \themodel{} for 500k steps using 16
H100 GPU nodes, with a batch size of 4 per GPU.
The learning rate peaks at $10^{-4}$ for both
encoder and decoder after 1k warmup steps,
followed by cosine decay to $10^{-6}$. For the training objective in
Eq.~\ref{eq:loss}, we set
$\lambda_{\mathrm{2D}}=
\lambda_{\mathrm{vis}}=
\lambda_{\mathrm{motion}}=\lambda_{\mathrm{reproj}}=0.1$,
$\lambda_{\mathrm{conf}}=0.2$, and
$\lambda_{\mathrm{normal}}=0.5$.
The resulting checkpoint is used for all
main evaluations and supports every input
configuration with a single set of weights,
without any additional test-time optimization.

\begin{table*}[t]
\centering
\small
\setlength{\tabcolsep}{10pt}
\resizebox{\linewidth}{!}{%
\renewcommand{\arraystretch}{0.88}%
\begin{tabular}{@{}l ccc ccc ccc ccc@{}}
\toprule
\multirow{2}[2]{*}{\textbf{Method}} & \multicolumn{3}{c}{\textbf{ADT}~\cite{pan2023adt}} & \multicolumn{3}{c}{\textbf{HOT3D}~\cite{banerjee2024hot3d}} & \multicolumn{3}{c}{\textbf{DexYCB}~\cite{chao2021dexycb}} & \multicolumn{3}{c}{\textbf{DROID}~\cite{khazatsky2024droid}} \\
\cmidrule(lr){2-4}\cmidrule(lr){5-7}\cmidrule(lr){8-10}\cmidrule(lr){11-13}
 & AJ~$\uparrow$ & $\mathrm{APD}_{3D}$~$\uparrow$ & OA~$\uparrow$
 & AJ~$\uparrow$ & $\mathrm{APD}_{3D}$~$\uparrow$ & OA~$\uparrow$
 & AJ~$\uparrow$ & $\mathrm{APD}_{3D}$~$\uparrow$ & OA~$\uparrow$
 & AJ~$\uparrow$ & $\mathrm{APD}_{3D}$~$\uparrow$ & OA~$\uparrow$ \\
\midrule
DELTA$^{\dagger}$~\cite{ngo2024delta} & 47.93 & 63.54 & 92.23 & 40.36 & 56.34 & 87.14 & 75.09 & 88.71 & 88.22 & 30.66 & 63.37 & 68.37 \\
TAPIP3D (RGBD)$^{\dagger}$~\cite{tapip3d} & 49.41 & 64.33 & 93.61 & 37.79 & 54.33 & 86.17 & 72.59 & 87.73 & 87.47 & 32.99 & 68.26 & 68.82 \\
Point4D$^{\ddagger}$~\cite{jeon2026point4d} & 59.66 & 77.56 & 88.09 & 9.53 & 16.25 & 85.48 & \third 78.08 & \third 92.68 & 88.66 & 31.94 & 70.78 & 59.08 \\
AllTracker $^{\dagger}$~\cite{harley2025alltracker} & 49.35 & 63.80 & \third 95.02 & 38.61 & 54.29 & \first 89.42 & 73.29 & 87.73 & 90.38 & 35.41 & 68.71 & 67.09 \\
CoTracker3 $^{\dagger}$~\cite{karaev25cotracker3} & 49.27 & 62.45 & \first 95.16 & 38.24 & 51.42 & 88.97 & 76.01 & 86.45 & \third 91.59 & 34.46 & 53.30 & \third 75.63 \\
Track4World$^{\S}$~\cite{lu2026track4world} & 55.78 & 69.51 & \second 95.14 & 9.39 & 16.07 & 88.65 & 76.02 & 88.53 & 91.49 & 32.25 & 66.31 & 70.94 \\
4RC$^{\ddagger}$~\cite{luo20264rc} & 46.90 & 67.14 & 79.04 & 25.83 & 41.04 & 79.70 & 73.42 & \second 93.32 & 82.49 & 20.26 & 53.84 & 40.02 \\
OmniX$^{\ddagger}$~\cite{jiang2026omnix} & 53.46 & 74.75 & 82.91 & 7.42 & 12.93 & 83.28 & 62.25 & 82.34 & 83.12 & 27.58 & 68.06 & 49.20 \\
SpaTracker-v2$^{\ddagger*}$~\cite{xiao2025spatialtrackerv2} & \first 67.12 & \second 80.40 & 94.86 & \third 47.38 & \third 64.84 & 88.80 & 73.79 & 87.91 & 89.33 & \third 36.03 & \third 73.52 & 72.48 \\
Any4D$^{\ddagger}$~\cite{karhade2025any4d} & 43.74 & 59.88 & 89.03 & 40.47 & 57.92 & 83.60 & 68.30 & 87.62 & 84.62 & 18.35 & 50.30 & 45.70 \\
OpenD4RT~\cite{zhang2026d4rt,opend4rt} & 31.04 & 44.21 & 90.81 & 7.98 & 16.86 & 77.21 & 70.51 & 84.97 & 90.00 & 28.43 & 58.91 & 67.19 \\
\cmidrule(l){1-13}
\textbf{Ours} (cam off) & \third 65.74 & \third 80.03 & 94.57 & \first 58.01 & \first 74.89 & \second 89.22 & \second 81.66 & 92.63 & \first 92.42 & \first 63.63 & \second 97.73 & \first 83.43 \\
\textbf{Ours} (cam on) & \second 67.08 & \first 81.18 & 94.67 & \second 52.72 & \second 70.86 & \third 89.03 & \first 83.33 & \first 93.98 & \second 92.19 & \second 63.44 & \first 97.86 & \second 83.34 \\
\bottomrule
\end{tabular}%
}
\caption{\textbf{Scale-aligned single-view 3D tracking.}
One median scale is fitted to ground truth per clip; visibility is unchanged.
Camera off disables MCPE while retaining intrinsics for UV-depth lifting.
Symbols and colors follow Table~\ref{tab:main_tracking}.}
\label{tab:scale_aligned_tracking}

\end{table*}

\subsection{3D Tracking Evaluation}

\noindent\textbf{Evaluation protocol.}
We evaluate metric 3D tracking on ADT~\cite{pan2023adt},
HOT3D~\cite{banerjee2024hot3d},
DexYCB~\cite{chao2021dexycb}, and
DROID~\cite{khazatsky2024droid}, with 50 clips
each from ADT, HOT3D, and DROID.
Following TAPVid-3D~\cite{koppula2024tapvid3d},
we report 3D Average Jaccard (AJ),
Average Percentage of Points within Delta
($\mathrm{APD}_{3D}$), and Occlusion Accuracy (OA).
AJ and $\mathrm{APD}_{3D}$ average over fixed metric
thresholds of 5, 10, 20, and 40\,cm,
while OA measures visibility prediction accuracy.
Unless otherwise stated, predictions are evaluated
directly in metric space without ground-truth
scale alignment.

ADT follows the benchmark query times and target-view
pixels~\cite{koppula2024tapvid}.
DexYCB uses the published query times and
target-camera-visible anchors.
Its 24--25-frame clips are repeat-padded to 48 frames
for methods requiring fixed-length inputs, with padded
frames excluded from evaluation.
HOT3D and DROID use fixed, visually diverse 48-frame subsets.
HOT3D trajectories track annotated hand and object mesh
vertices, with visibility determined by mesh occlusion,
calibrated image bounds, and arm/hand segmentation masks
in rectified pinhole views.
DROID trajectories track sampled robot surfaces using
recorded forward kinematics, with visibility determined
by mesh occlusion and SAM3~\cite{carion2025sam3} foreground masks.
Queries begin at the first valid observation in the
reference camera (HOT3D RGB and DROID exterior cam1),
with at most 512 queries per DROID clip.
Both datasets use identical frozen anchor queries
across single- and multi-view evaluation.
Ground-truth annotations define query locations and
evaluation trajectories, but ground-truth depth is
never provided as a model input.
$\mathrm{APD}_{3D}$ is evaluated on ground-truth-visible
points, and clips are weighted equally within each dataset.\looseness=-1 

Our model receives camera information only through MCPE.
For evaluation, intrinsics are used to lift predicted
image positions and depth into 3D, while benchmark camera
poses transform predictions into a common coordinate frame,
including when MCPE is disabled.
These transformations are applied only during evaluation
and do not provide camera information to the model
when MCPE is disabled.

Baselines requiring metric depth use DA3-Metric,
DA3-Nested~\cite{depthanything3},
\mbox{MoGe-3}~\cite{kong2026moge3}, or
UniDepth-v2~\cite{piccinelli2025unidepthv2},
all given known intrinsics.
Depth-assisted baselines use DA3-Nested, which
additionally receives metric poses for moving-camera
sequences, unless noted.
The symbols in Tables~\ref{tab:mv_tracking} and
\ref{tab:main_tracking} indicate the use of external
depth, while Table~\ref{tab:tracking_depth_inputs}
summarizes the inputs available to each depth provider.
Table~\ref{tab:single_view_depth_variants} in the appendix repeats the
single-view comparison with each of the four depth providers.
TAPIP3D RGBD additionally receives known intrinsics,
poses, and estimated depth.
DELTA and the AllTracker and CoTracker3 lifting
pipelines use known intrinsics, while Point4D
and OmniX estimate camera geometry.
On HOT3D, our evaluation adapter supplies known
intrinsics and poses to the DA3 component of
Track4World, although these inputs are not supported
by its released interface.
Tables~\ref{tab:main_tracking} and
\ref{tab:scale_aligned_tracking} also include
OpenD4RT~\cite{opend4rt}, an unofficial implementation
of D4RT~\cite{zhang2026d4rt}, evaluated using its
public 48-frame checkpoint
\texttt{OpenD4RT\_48CLIP\_9Mix\_NoCropAUG}.

\noindent\textbf{Multi-view metric results.}
With three calibrated views and no external depth,
\themodel achieves the highest metric AJ on all four
benchmarks against every tracker and depth provider
(Table~\ref{tab:mv_tracking}), with the largest margins
on HOT3D (+16.4 AJ) and DROID (+41.4 AJ) over the strongest baseline.

\noindent\textbf{Single-view metric results.}
With the same weights and a single calibrated video,
\themodel also achieves the highest AJ on all four
benchmarks (Table~\ref{tab:main_tracking}), dropping
less than 3 AJ points from the three-view setting on
HOT3D and DROID. The margin is smallest on DexYCB
(1.9 AJ), whose static cameras provide no motion cue
for scale.

\noindent\textbf{Scale-aligned results.}
Metric tracking performance reflects both global
scale accuracy and residual 3D reconstruction
and tracking errors.
Table~\ref{tab:scale_aligned_tracking} evaluates
performance after fitting a single median scale
factor per clip.
This corrects global scale differences but does
not eliminate local geometric distortions,
trajectory errors, or differences in method inputs.
With or without camera conditioning, \themodel
outperforms all baselines in AJ on HOT3D, DexYCB,
and DROID. On ADT, it matches SpaTracker-v2,
which uses test-time optimization, in AJ.
Camera-conditioned \themodel achieves the highest
$\mathrm{APD}_{3D}$ across all four benchmarks,
indicating that its advantages extend beyond
metric scale estimation to relative 3D geometry
and motion.

\begin{figure*}[t]
    \centering
    \includegraphics[width=\textwidth]{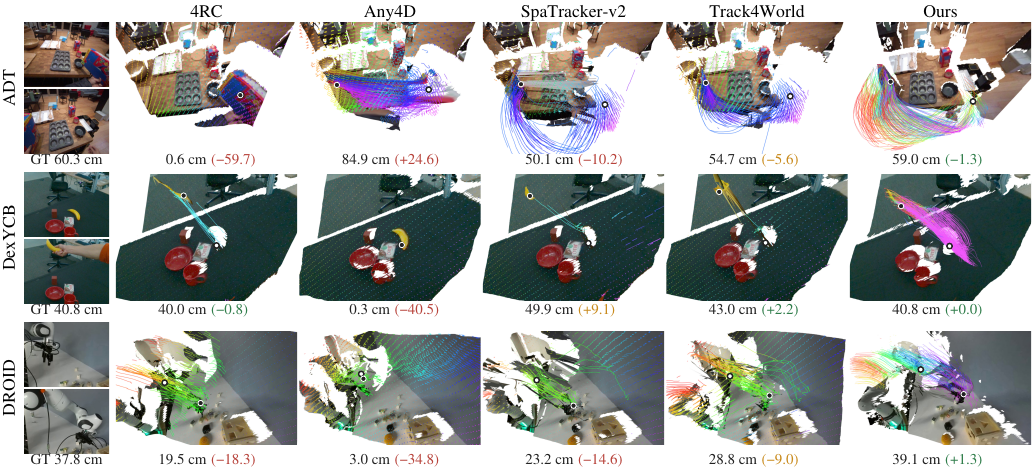}
    \caption{\textbf{Metric 4D tracking results across methods.}
    Each row shows the first and last input frames of a clip and its last frame from one viewpoint near the input camera, shared by all methods.
    Below each panel: the moved object's travel from the first to the last frame and its error in \text{cm}, with ground truth (GT) under the input frames.
    4RC, Any4D and SpaTracker-v2 take their metric scale from DA3-Metric~\cite{depthanything3}; Track4World uses its embedded DA3.}
    \label{fig:qual_tracking_main}
    \label{fig:qual_tracking}
\end{figure*}

\begin{table}[t]
\centering
\small
\setlength{\tabcolsep}{2pt}
\resizebox{\linewidth}{!}{%
\begin{tabular}{@{}lc ccc ccc@{}}
\toprule
\multirow{2}[2]{*}{\textbf{Method}} & \multirow{2}[2]{*}{\textbf{Cam.}} & \multicolumn{3}{c}{\textbf{ScanNet}~\cite{dai2017scannet}} & \multicolumn{3}{c}{\textbf{Bonn}~\cite{palazzolo2019refusion}} \\
\cmidrule(lr){3-5}\cmidrule(lr){6-8}
 & & AbsRel~$\downarrow$ & $\delta_{1.25}$~$\uparrow$ & TAE~$\downarrow$ & AbsRel~$\downarrow$ & $\delta_{1.25}$~$\uparrow$ & TAE~$\downarrow$ \\
\midrule
DA3\textsc{nested}~\cite{depthanything3} & off & \second 0.062 & \second 0.970 & \third 1.00 & \third 0.137 & \second 0.902 & \third 2.43 \\
MapAnything~\cite{keetha2026mapanything} & off & 0.317 & 0.450 & 1.14 & 0.218 & 0.760 & 2.62 \\
Any4D~\cite{karhade2025any4d} & off & 0.340 & 0.233 & 1.24 & 0.445 & 0.018 & 2.65 \\
\midrule
\textbf{Ours} & off & \third 0.083 & \third 0.945 & \second 0.68 & \second 0.133 & \third 0.895 & \first 1.76 \\
\textbf{Ours} & on & \first 0.041 & \first 0.979 & \first 0.62 & \first 0.107 & \first 0.967 & \second 1.80 \\
\bottomrule
\end{tabular}%
}
\caption{\textbf{Metric video depth.}
\text{Camera on} uses
known intrinsics, metric poses, and world up; off uses RGB only as input. No GT scale alignment. 
TAE is reported in percent.
}
\label{tab:main_depth}
\end{table}

\subsection{Dense Metric Depth}
\label{sec:exp_depth}

Beyond 3D tracking, \themodel directly predicts dense metric depth
without task-specific fine-tuning.
We evaluate on ScanNet~\cite{dai2017scannet} and
Bonn~\cite{palazzolo2019refusion}
using AbsRel, $\delta_{1.25}$, and temporal alignment error
(TAE)~\cite{yang2025depthanyvideo}, which measures depth
disagreement between adjacent frames after pose-based warping.
All metrics are computed on the first 48 frames of each sequence
without ground-truth scale or shift alignment.

Camera inputs for each method are summarized
in Table~\ref{tab:main_depth}.
We exclude two ScanNet sequences
due to inaccurate camera poses.
DA3-Nested uses its unconditioned video-depth interface.
TAE uses ground-truth depth only to define validity masks.
For Bonn, both of our configurations use the published
marker-to-optical-camera transformation.

\noindent\textbf{Results.}
As shown in Table~\ref{tab:main_depth},
RGB-only inference outperforms MapAnything and Any4D
in both AbsRel and $\delta_{1.25}$.
Camera conditioning further improves depth accuracy,
achieving the best AbsRel and $\delta_{1.25}$
on both datasets.
Both configurations also outperform all baselines
in TAE, showing strong temporal consistency of the predicted depth.

\subsection{Qualitative Results}
\label{sec:exp_qualitative}
\begin{figure*}[p]
\centering
\includegraphics[width=0.96\textwidth]{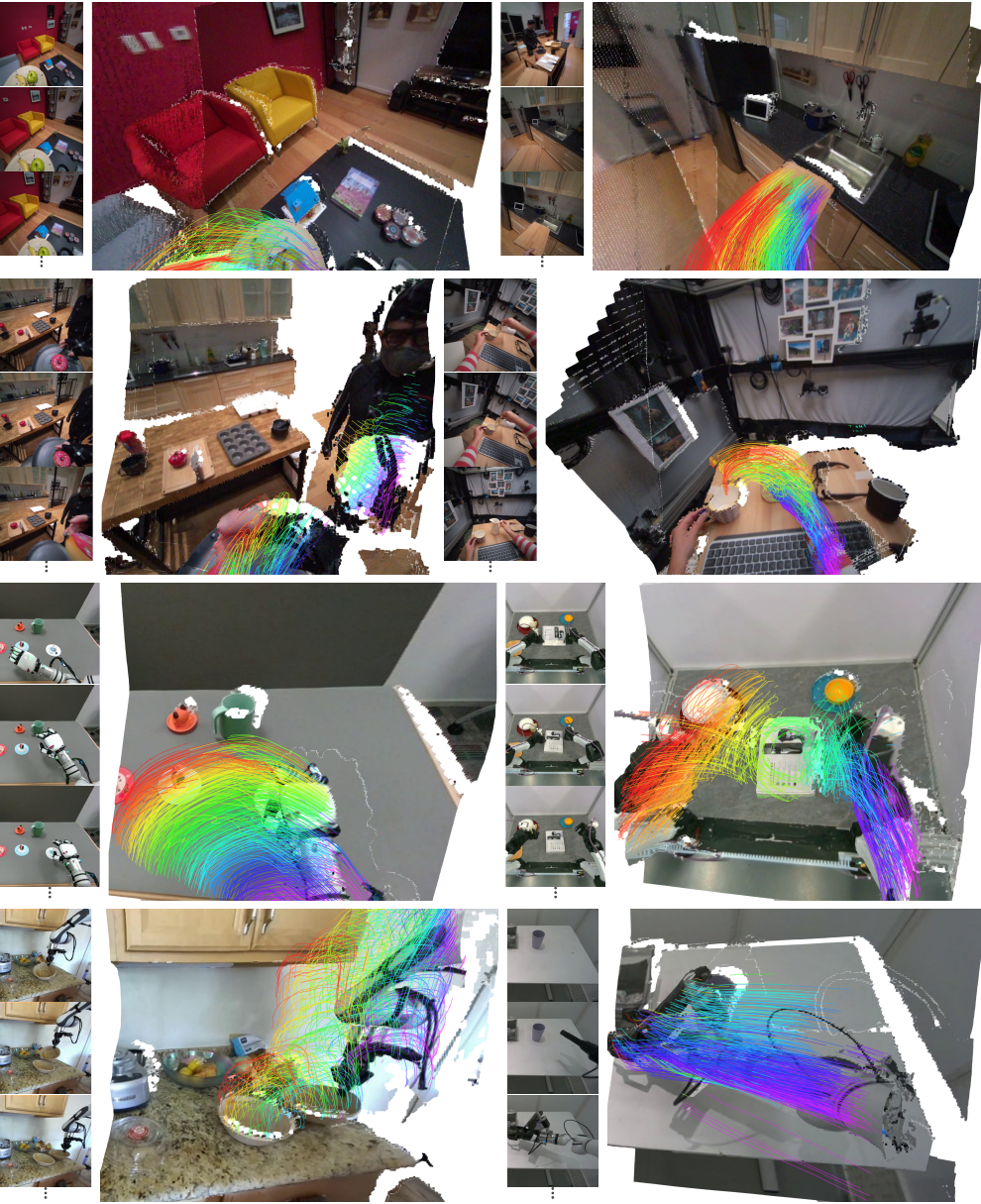}
\ifmetareport
\caption{\textbf{Qualitative results on egocentric and robot videos.}
Each example shows the first, middle, and last input frames and our
prediction at the last frame: the reconstructed surface and the
trajectories of moving points, colored by pixel position, seen slightly
off the input camera. Rows 1--2: egocentric ADT and HOT3D videos with
known cameras (one view; three for the right ADT example in row 1).
Rows 3--4: robot manipulation; VITRA, ABC, and RealSource use one RGB
view without camera input, and DROID uses two calibrated views.}
\else
\caption{\textbf{Additional results of \themodel.}
Each example shows the first, middle, and last input frames and our
prediction at the last frame: the reconstructed surface and the
trajectories of moving points, colored by pixel position, seen slightly
off the input camera. Rows 1--2 show egocentric videos from ADT and HOT3D,
and rows 3--4 show robot manipulation.}
\fi
\label{fig:supp_ours}
\end{figure*}

\begin{figure*}[p]
\centering
\includegraphics[width=0.935\textwidth]{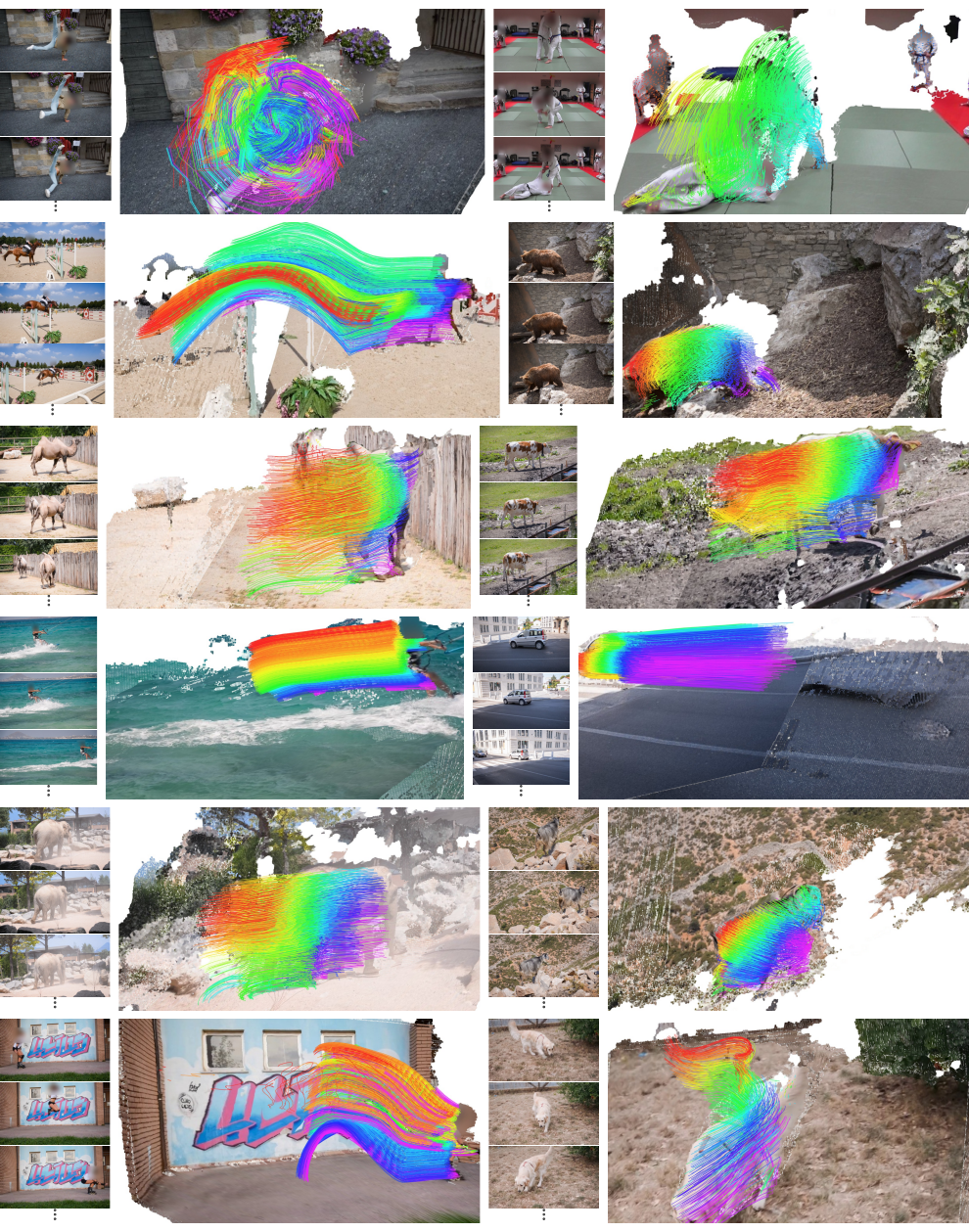}
\caption{\textbf{Results on DAVIS.}
One RGB video with no camera input; \themodel estimates the focal length
and camera motion. Each example shows the first, middle, and last input
frames and our prediction, rendered as in Figure~\ref{fig:supp_ours}.
Faces are blurred.}
\label{fig:supp_davis}
\end{figure*}

Fig.~\ref{fig:qual_tracking_main} compares metric 4D tracking
across methods from shared novel viewpoints. Baselines exhibit
static-point drift and ray-like depth errors, while \themodel
maintains coherent geometry and motion, with object displacement
errors within 1.5\,cm across all three examples.
Fig.~\ref{fig:supp_ours} and~\ref{fig:supp_davis} further demonstrate \themodel
on diverse real-world scenarios, including dynamic outdoor
scenes, human-object interactions, and robotic manipulation.
Across these settings, \themodel reconstructs coherent 3D
geometry and tracks fast and complex motion.
Fig.~\ref{fig:in_the_wild} in Appendix~\ref{app:in_the_wild} shows
eight more in-the-wild recordings.

\subsection{Camera-Asymmetric Self-Improvement}
\label{sec:exp_flywheel}
\label{sec:exp_hoi}
\label{sec:exp_data_scale}

\begin{figure}[t]
\centering
\includegraphics[width=\linewidth]{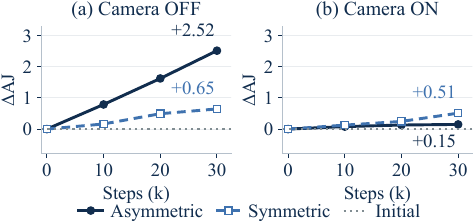}\par
\vspace{-5pt}
\caption{\textbf{Camera-Asymmetric Self-Improvement.}
AJ gain over initialization ($\Delta$AJ, no scale alignment), averaged
over ADT and HOT3D with three views and EMA weights, under
(a) camera-off and (b) camera-on inference. The teacher is
camera-conditioned (asymmetric) or camera-off (symmetric).}
\label{fig:hoi_self_distillation}
\end{figure}

We investigate whether \themodel can leverage its
stronger camera-conditioned predictions to improve
RGB-only inference through self-training
(Fig.~\ref{fig:hoi_self_distillation}).
We evaluate on the egocentric ADT and HOT3D
benchmarks, starting all runs from a shared checkpoint.\looseness=-1

\begin{figure}[t]
\centering
\includegraphics[width=\linewidth]{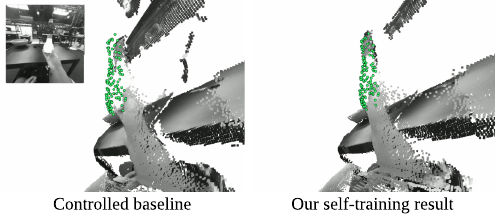}
\caption{\textbf{Camera-asymmetric self-training visualization.}
Left: symmetric controlled baseline; right: our camera-asymmetric self-training result. Green points mark ground truth.}
\label{fig:self_training_ranch}
\end{figure}

\noindent\textbf{Self-training setup.}
As a control, symmetric self-training uses the
same recipe, including initialization, unlabeled
videos, replay, and EMA teacher, but disables
camera conditioning in the teacher, so teacher
and student receive identical inputs.
This comparison isolates the effect of camera
conditioning in the teacher, the only difference
between the two runs. Appendix~\ref{app:arxiv_details} gives the
fine-tuning details.

\noindent\textbf{Results.}
A camera-conditioned teacher provides more effective
supervision for a camera-free student than symmetric
camera-off self-training.
With camera-off inference, the advantage grows during
training (Fig.~\ref{fig:hoi_self_distillation}a),
yielding a mean AJ improvement of 2.52 points
over initialization after 30k steps, compared with
0.65 points for symmetric self-training.
Between 20k and 30k steps, symmetric self-training
adds only 0.2 AJ, while the asymmetric student
adds 0.9 AJ.
Improvements on both ADT (+1.29 AJ) and
HOT3D (+3.75 AJ) demonstrate the benefit of
transferring calibration-informed geometric
knowledge to RGB-only prediction.
These gains do not come at the expense of
camera-conditioned performance, which stays
within 0.15 AJ of the initial model
(Fig.~\ref{fig:hoi_self_distillation}b).
Fig.~\ref{fig:self_training_ranch} shows a qualitative example on a
Show3D clip, where both students run on RGB only and see the same frames;
green points mark the ground-truth positions
of the object.

\section{Design Ablations}
\label{app:additional_ablations}
\suppressfloats[t]

\subsection{Geometric Conditioning and Metric Scale}
\label{app:camera_encoding}
\label{sec:exp_camera}
\label{app:metric_ablation}
\label{app:inference_inputs}

\begin{figure}[!t]
\centering
\includegraphics[width=\linewidth]{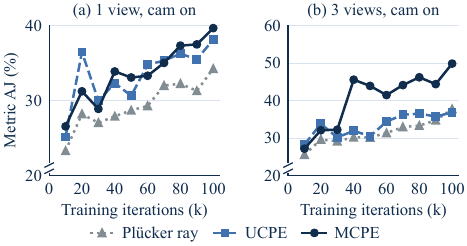}
\caption{\textbf{Camera encoding during training.}
Panels (a,b) use one and three input views, respectively.
Cameras are on at evaluation; all runs train with 50\%
camera conditioning.}
\label{fig:camera_encoding_training_curves}
\end{figure}

\begin{figure}[!t]
\centering
\includegraphics[width=\linewidth]{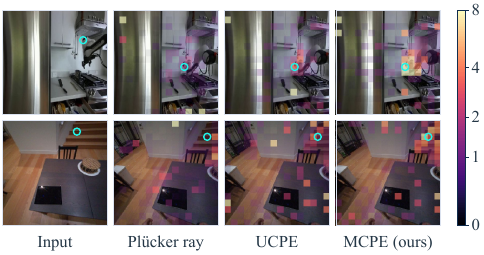}

\caption{\textbf{Attention under different camera encodings.}
Block-40 attention maps for two queries using Pl\"ucker-ray
embeddings, UCPE, and MCPE, with cameras enabled.
Cyan circles mark queries and ground-truth correspondences.
Attention is averaged over 16 heads and normalized by
uniform attention (logarithmic color scale).}
\label{fig:camera_conditioning}
\label{fig:camera_attention}
\end{figure}

\noindent\textbf{Encoding approaches.}
We compare additive Pl\"ucker-ray embeddings with
geometric attention (UCPE and MCPE) across different
view counts and camera-conditioning settings.
Fig.~\ref{fig:camera_encoding_training_curves}
shows that all three approaches achieve comparable
performance without camera conditioning, while
geometric attention, particularly MCPE, provides
substantially higher metric AJ when cameras are available.
The advantage is especially pronounced with three views,
highlighting MCPE's ability to exploit calibrated geometry
without compromising RGB-only inference.
This motivates our design of separate RGB and camera
attention branches for flexible camera conditioning.
The attention maps in Fig.~\ref{fig:camera_attention}
further illustrate how the different encodings attend
to ground-truth correspondences.\looseness=-1

\noindent\textbf{Dual-frame rays under camera rotation.}
\begin{figure}[t]
\centering
\includegraphics[width=\linewidth]{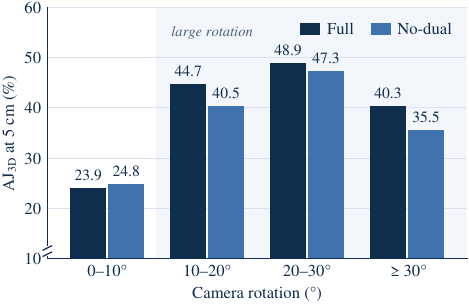}
\caption{
\textbf{Dual-frame rays under camera rotation.}
Whole-clip AJ at 5\,cm, grouped by the maximum
within-pair camera rotation across all views.
Full outperforms No-dual on clips with maximum
rotations of at least $10^\circ$, with the largest
gain at $\geq 30^\circ$ (+4.8 AJ).
The shaded region indicates rotations of at least
$10^\circ$.
}
\label{fig:dual_frame_rotation}
\end{figure}

We compare the Full and No-dual models after
50k fine-tuning updates from a shared initialization,
using three views with camera conditioning.
All clips are grouped by the maximum camera
rotation within the original encoder frame pairs
$(0,1),(2,3),\ldots$, across all three views.
We preserve the original frame sampling and queries
and evaluate whole-clip AJ at a 5\,cm threshold,
weighting clips equally within each bin.
As shown in Fig.~\ref{fig:dual_frame_rotation},
Full improves tracking on clips with within-pair
rotations of at least $10^\circ$, with the largest
gain at $\geq 30^\circ$ (40.30 vs.\ 35.54 AJ).
This suggests that representing both temporal ray
frames is particularly beneficial when camera
rotation produces substantial changes in viewing
direction within an encoder token.\looseness=-1

\begin{figure}[t]
\centering
\includegraphics[width=\linewidth]{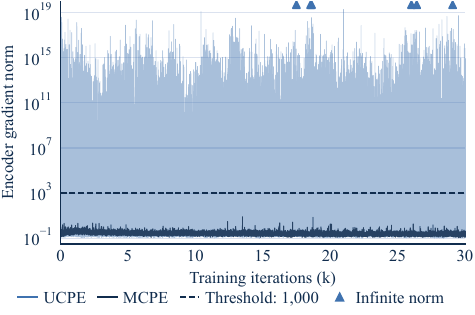}
\caption{\textbf{Encoder gradient norms during training.}
Raw pre-clipping norms on a logarithmic axis.
The dashed line marks $10^3$, triangles denote infinite
UCPE norms.}
\label{fig:encoder_gradient_norm}
\end{figure}

\noindent\textbf{Training stability.}
Fig.~\ref{fig:encoder_gradient_norm} compares raw encoder
gradient norms over aligned 30k-iteration windows.
Uncapped UCPE exhibits frequent large gradient spikes
and occasional infinite norms, whereas MCPE maintains
substantially smaller and more stable gradients.
This comparison illustrates the improved numerical
stability of span normalization and scale conditioning.\looseness=-1

\phantomsection
\noindent\textbf{Learning metric scale from cameras.}
\label{sec:exp_metric_space}
Fig.~\ref{fig:camera_scale_response} examines whether
\themodel uses camera baselines as a metric scale cue.
With RGB observations and queries fixed, we rescale camera
translations by $\alpha\in\{0.8,1.0,1.2\}$ and examine the
resulting geometry (Fig.~\ref{fig:camera_scale_response}a).
We quantify scale sensitivity by the ratio
$R_{\mathrm{scale}}$ of mean predicted depths at
$\alpha=1.2$ and $\alpha=0.8$.
A proportional response yields $R_{\mathrm{scale}}=1.5$,
whereas scale invariance yields 1.
Averaged equally across four benchmarks,
scale augmentation increases $R_{\mathrm{scale}}$
from 1.069 to 1.393, bringing the response closer
to the expected proportional scaling (Fig.~\ref{fig:camera_scale_response}b).
Scale augmentation also improves AJ at a 5\,cm threshold
from 22.94 to 25.51 (Fig.~\ref{fig:camera_scale_response}c). 

\begin{figure}[t]
\input{figures/fig_camera_scale_response}
\end{figure}

\begin{figure}[t]
\centering
\captionsetup{skip=2pt}
\includegraphics[width=\linewidth]{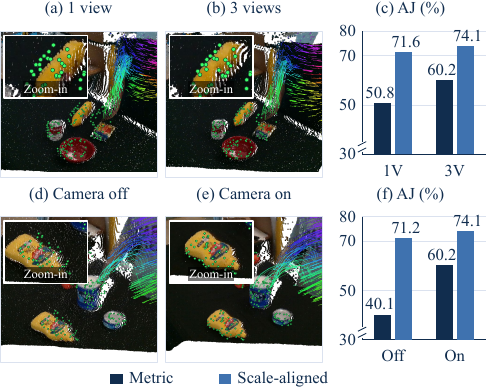}
\caption{\textbf{Benefits of Geometric Measurements.}
Additional views (a-c) and camera conditioning (d-f) improve metric 3D tracking.
Green markers indicate ground truth.}
\label{fig:inference_view_scaling}
\label{fig:qualitative_input_views}
\label{fig:qualitative_camera_conditioning}
\end{figure}

\phantomsection
\noindent\textbf{Geometric Measurements.}
\label{sec:exp_geometric_inputs}
Fig.~\ref{fig:inference_view_scaling} evaluates the benefits of additional measurements at inference using identical model weights. Increasing input views from one to three improves metric AJ by 9.4 points and scale-aligned AJ by 2.4 points, indicating improved 3D localization beyond scale correction. Enabling camera conditioning improves metric AJ by 20.1 points but scale-aligned AJ by only 2.8 points, highlighting the role of calibrated cameras in recovering metric scale. Qualitative results further illustrate improved geometry and trajectory accuracy with richer geometric inputs.

\begin{figure}[t]
\input{figures/fig_zero_shot_view_count}
\end{figure}

\phantomsection
\label{app:input_view_count}

\begin{figure*}[t]
\input{figures/fig_zero_shot_view_qualitative}
\end{figure*}

\subsection{Zero-Shot Inference Beyond Three Views}
\label{app:zero_shot_view_count}

Although \themodel is trained with at most three views,
its architecture supports a variable number of inputs.
We evaluate zero-shot generalization to up to ten views
using the frozen checkpoint with camera conditioning
on PStudio~\cite{joo2015panoptic} and
Show3D~\cite{rim2026show3d}.

\noindent\textbf{Inference extension.}
The encoder processes a variable number of views
without learned view-index embeddings.
For additional views, we extend the runtime view-count
limits and reuse the decoder's learned view embeddings,
keeping the source-view embedding fixed and alternating
the two auxiliary-view embeddings.
All selected views are jointly encoded, and the decoder
attends to their combined features without changing
model weights or performing any additional test-time optimization.

\noindent\textbf{Evaluation protocol.}
We evaluate 20 clips each from PStudio and Show3D,
using the same frozen clips and queries across methods
and view counts.
The target camera remains in slot zero.
Each setting uses the first $N$ cameras from a fixed
ten-camera list: the target followed by its nine
nearest camera centers for PStudio, and cam1--cam10
for Show3D.
For \themodel, we use 48 frames at $224\times224$
resolution and identical target-camera queries
at every view count from one to ten.

\noindent\textbf{Multi-view baselines.}
We compare against released checkpoints of
MVTracker~\cite{rajic2025mvtracker},
ARROW~\cite{fradlin2026arrow}, and
OmniX~\cite{jiang2026omnix}, using identical clips,
camera subsets, and queries at one through ten views.
MVTracker receives RGB, calibration, and DA3-Nested
metric depth~\cite{depthanything3}.
ARROW and OmniX estimate camera geometry from RGB
and recover metric scale using DA3-Nested
target-view depth.
All methods retain their native image preprocessing.
For OmniX, we extend the view-count limit without
changing model weights and restrict motion queries
to target-camera source frames and output times,
while retaining all camera features for cross-attention.
This enables 48-frame inference within 80\,GB
without reducing image resolution.
A three-view, four-frame check reproduces the
original predictions exactly.
Depth estimates are reused across view counts,
and no method uses ground-truth depth or
trajectories for scale fitting.

\noindent\textbf{Results.}
Fig.~\ref{fig:zero_shot_view_count} shows that
\themodel improves consistently from one to three
views and continues to benefit from additional
views beyond its training limit.
From three to ten views, PStudio gains 22.91 AJ
and 25.29 $\mathrm{APD}_{3D}$ points, while
Show3D gains 3.60 AJ and 3.86 $\mathrm{APD}_{3D}$
points.
At ten views, \themodel outperforms all three
baselines on both datasets in metric AJ and
$\mathrm{APD}_{3D}$.
The baselines show limited improvement with
additional views on PStudio; on Show3D, ARROW
degrades while MVTracker and OmniX remain
nearly unchanged.
These results demonstrate that \themodel can
exploit additional calibrated observations
beyond the view counts encountered during
training, without updating its weights.
Fig.~\ref{fig:zero_shot_view_qualitative}
shows two qualitative examples.\looseness=-1

\subsection{Training Recipe Ablations}
\label{app:recipe_comparison}
\label{app:dataset_mixture_ablation}
\label{app:training_conditioning_frequency}

\begin{figure}[t]
\centering
\includegraphics[width=\linewidth]{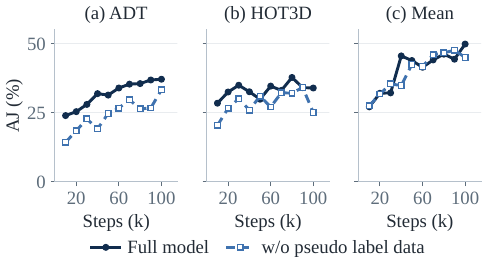}\par
\vspace{-5pt}
\caption{\textbf{Ablation on pseudo-labeled training data.}
Three-view, camera-conditioned metric AJ with and without
pseudo-labeled training data.
(a) ADT, (b) HOT3D, and (c) the equally weighted mean
across ADT, HOT3D, DexYCB, and DROID. \vspace{-2mm}}
\label{fig:dataset_mixture_ablation}
\end{figure}

\noindent\textbf{Pseudo-labeled training data.}
Our training mixture includes egocentric stereo
pseudo-labeled clips (Sec.~\ref{sec:data_labeling}).
Fig.~\ref{fig:dataset_mixture_ablation} compares
the full mixture with a variant excluding these clips,
while keeping the remaining training recipe unchanged.
At 100k iterations, the full mixture improves metric AJ
from 33.32 to 37.14 on ADT and from 25.15 to 33.97
on HOT3D, while the four-dataset mean remains
broadly comparable across checkpoints.
These results suggest that egocentric pseudo-labeled
data particularly benefits benchmarks with similar
visual domains, without substantially changing
overall performance.

\begin{figure}[t]
\centering
\includegraphics[width=\linewidth]{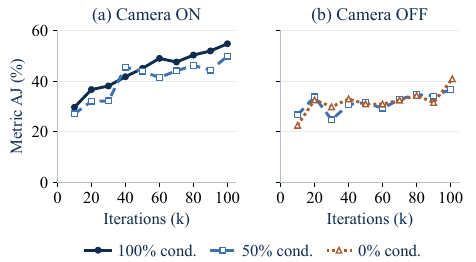}\par
\vspace{-5pt}
\caption{\textbf{Ablation on camera-conditioning frequency.}
Three-view metric AJ averaged over all test clips.
(a) Camera-on inference for models trained with
50\% or 100\% camera conditioning.
(b) Camera-off inference for models trained with
0\%, 50\%, or 100\% camera conditioning.\vspace{-2mm}}
\label{fig:training_conditioning_frequency}
\end{figure}

\noindent\textbf{Camera-conditioning frequency.}
Fig.~\ref{fig:training_conditioning_frequency}
compares models trained with camera conditioning
enabled for 0\%, 50\%, or 100\% of examples.
With cameras enabled at inference, 100\% conditioning
achieves the highest AJ, while 50\% conditioning
remains competitive.
With cameras disabled, 50\% conditioning achieves
36.73 AJ at 100k iterations, comparable to the
0\%-conditioned model.
These results show that camera dropout enables
a single model to support both inference settings,
balancing camera-conditioned accuracy with
camera-free performance.
We use 50\% conditioning for the final model
(Sec.~\ref{sec:final_recipe}).

\section{Conclusion}

We introduce \themodel, a unified feedforward model for metric
4D perception across diverse sensing configurations.
By effectively leveraging geometric measurements, \themodel
delivers accurate metric geometry and motion predictions
with flexible camera conditioning.
Despite being trained with at most three views, \themodel
generalizes zero-shot to larger numbers of views and continues
to benefit from additional observations at inference.
We further demonstrate that its flexible camera conditioning
enables asymmetric self-training, where camera-conditioned
predictions improve RGB-only inference on unlabeled videos.
Together, these results establish \themodel as a scalable and
flexible foundation for metric 4D perception and future
self-training research on unlabeled calibrated videos.\looseness=-1

{
    \small
    \setlength{\bibsep}{4.75pt plus 0.5pt}
    \bibliographystyle{cvpr2027/ieeenat_fullname}
    \bibliography{vedaldi_general_pruned,vedaldi_specific,scale4d}
}

\twocolumn[{%
\appendix{\huge\sffamily\bfseries Appendix\par}
\vspace{10pt}
}]
\renewcommand{\dbltopfraction}{0.95}
\makeatletter\setlength{\@dblfptop}{0pt}\setlength{\@fptop}{0pt}\makeatother   %

\section{Training Data}
\label{app:training_sources}
Table~\ref{tab:flywheel_data} summarizes the public training
sources of Sec.~\ref{sec:data}: the number of training samples
and video frames contributed by each dataset. Sources with
uncertain scale are rescaled and filtered as described there.
Our internal stereo clips, labeled with the pipeline of
Sec.~\ref{sec:data_labeling}, are not listed.

\section{Experimental Details}
\label{app:arxiv_details}

\noindent\textbf{Camera-asymmetric self-training (Sec.~\ref{sec:exp_flywheel}).}
A camera-conditioned teacher supervises a
camera-free student using unlabeled calibrated
videos with labeled replay.
Fine-tuning uses 2 H100 GPU nodes with a global
batch size of 64.
We use cosine learning-rate schedules from
$10^{-6}$ to $2\times10^{-7}$ for the encoder
and from $10^{-5}$ to $2\times10^{-6}$ for
other parameters.

\noindent\textbf{Multi-view baselines (Sec.~\ref{app:zero_shot_view_count}).}
MVTracker uses RGB, calibration, and DA3-Nested
metric depth~\cite{depthanything3}.
ARROW and OmniX estimate camera geometry from RGB,
with metric scale recovered using DA3-Nested
target-view depth.
All methods retain their native image preprocessing.
For OmniX, we extend the view-count limit without
changing model weights and restrict motion queries
to target-camera source frames and output times,
while retaining all camera features for cross-attention.
This enables 48-frame inference within 80\,GB
without reducing image resolution.
A three-view, four-frame check reproduces the
original predictions exactly.
Depth estimates are cached and reused across
view counts, and no method uses ground-truth
depth or trajectories for scale fitting.\looseness=-1

\noindent\textbf{Robustness to depth providers.}
To assess sensitivity to external depth estimates,
Table~\ref{tab:single_view_depth_variants} compares
DA3-Nested with DA3-Metric, MoGe-3, and UniDepth-v2
for each applicable baseline, while our model
remains unchanged.
$\pi^4$ (native) uses its predicted scale without
depth-provider rescaling.
\themodel achieves the highest AJ on ADT, HOT3D,
and DROID across all tested depth providers.
On DexYCB, the strongest baseline configuration
uses MoGe-3 depth, indicating greater sensitivity
to the choice of depth provider.
Table~\ref{tab:tracking_depth_inputs} lists the camera
inputs that each depth provider receives.

\begin{table}[!htbp]
\centering
\footnotesize
\setlength{\tabcolsep}{3pt}
\begin{tabular*}{\columnwidth}{@{}l@{\extracolsep{\fill}}rr@{}}
\toprule
\textbf{Data source} & \textbf{Training samples} & \textbf{Video frames} \\
\midrule
\multicolumn{3}{@{}l}{\textit{Aggregated datasets with annotations}} \\
\addlinespace[2pt]
Dynamic Replica~\cite{karaev23dynamicstereo} & 483       & 144{,}900 \\
PointOdyssey~\cite{zheng2023pointodyssey}    & 8{,}191   & 1{,}048{,}448 \\
FallingThings~\cite{tremblay2018fallingthings}   & 1{,}013   & 14{,}120 \\
Kubric~\cite{greff2022kubric}          & 13{,}862  & 782{,}640 \\
CO3Dv2~\cite{reizenstein2021co3d}          & 5{,}905   & 1{,}182{,}963 \\
TartanAir~\cite{wang2020tartanair}       & 1{,}098   & 275{,}887 \\
GigaHands~\cite{fu2025gigahands}       & 2{,}818   & 589{,}399 \\
TartanGround~\cite{patel2025tartanground}    & 1{,}467   & 397{,}961 \\
Waymo Open~\cite{sun2020waymo}      & 732       & 145{,}012 \\
\bottomrule
\end{tabular*}
\caption{\textbf{Training data statistics.} Breakdown of the number of training samples and video frames across public datasets.}
\label{tab:flywheel_data}
\end{table}

\begin{table}[!ht]
\centering
\small
\setlength{\tabcolsep}{4pt}
\begin{tabular}{@{}l>{\raggedright\arraybackslash}p{0.3\linewidth}>{\raggedright\arraybackslash}p{0.3\linewidth}@{}}
\toprule
\textbf{Depth provider} & \textbf{Known intrinsics} & \textbf{Known metric poses} \\
\midrule
DA3-Metric~\cite{depthanything3} & Focal conversion to meters & Not supplied \\
DA3-Nested~\cite{depthanything3} & Network input & Supplied for translating camera streams \\
MoGe-3~\cite{kong2026moge3} & Horizontal field of view & Not supplied \\
UniDepth-v2~\cite{piccinelli2025unidepthv2} & Intrinsics matrix input & Not supplied \\
\bottomrule
\end{tabular}
\caption{\textbf{Camera inputs to the tracking depth providers}, in addition to each
tracker's own camera interface. DA3-Nested omits poses for static cameras such as DexYCB's.}
\label{tab:tracking_depth_inputs}
\end{table}

\newcommand{\appendixwildfigure}{%
  \begingroup   %
  \centering
  \includegraphics[width=\textwidth]{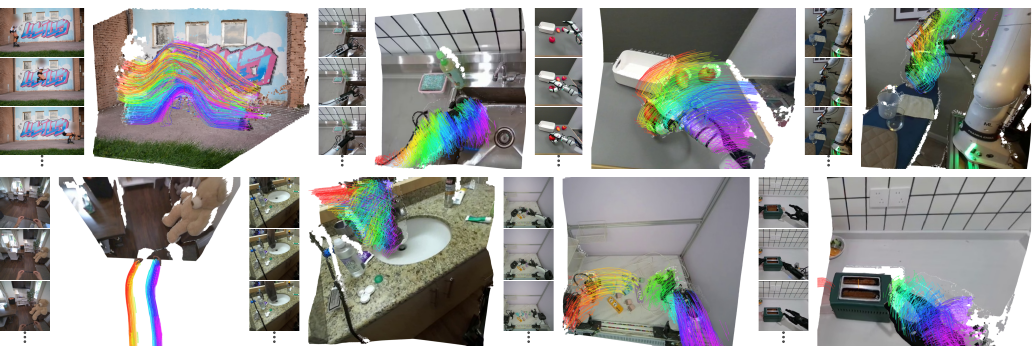}
  \captionof{figure}{\textbf{In-the-wild 4D metric perception.} Eight more recordings of people and robots, seen from novel viewpoints.}
  \label{fig:in_the_wild}
  \par\endgroup\vspace{4pt}}
\begin{table*}[tp]
\ifdefined\appendixwildfigure\appendixwildfigure\fi
\centering
\scriptsize
\setlength{\tabcolsep}{2.6pt}
\renewcommand{\arraystretch}{0.92}   %
\resizebox{\linewidth}{!}{%
\begin{tabular}{@{}ll ccc ccc ccc ccc@{}}
\toprule
\multirow{2}[2]{*}{\textbf{Method}} & \multirow{2}[2]{*}{\textbf{Depth model}} & \multicolumn{3}{c}{\textbf{ADT}~\cite{pan2023adt}} & \multicolumn{3}{c}{\textbf{HOT3D}~\cite{banerjee2024hot3d}} & \multicolumn{3}{c}{\textbf{DexYCB}~\cite{chao2021dexycb}} & \multicolumn{3}{c}{\textbf{DROID}~\cite{khazatsky2024droid}} \\
\cmidrule(lr){3-5} \cmidrule(lr){6-8} \cmidrule(lr){9-11} \cmidrule(lr){12-14}
 & & AJ~$\uparrow$ & $\mathrm{APD}_{3D}$~$\uparrow$ & OA~$\uparrow$ & AJ~$\uparrow$ & $\mathrm{APD}_{3D}$~$\uparrow$ & OA~$\uparrow$ & AJ~$\uparrow$ & $\mathrm{APD}_{3D}$~$\uparrow$ & OA~$\uparrow$ & AJ~$\uparrow$ & $\mathrm{APD}_{3D}$~$\uparrow$ & OA~$\uparrow$ \\
\midrule
\multirow{4}{*}{DELTA$^{\dagger}$~\cite{ngo2024delta}} & DA3-Metric~\cite{depthanything3} & 28.55 & 39.97 & 92.23 & 32.00 & 48.38 & 87.14 & 64.81 & 78.60 & 88.22 & 23.92 & 50.28 & 68.37 \\
 & DA3-Nested~\cite{depthanything3} & 42.62 & 52.88 & 93.81 & 22.09 & 34.57 & 87.54 & 63.81 & 74.51 & 90.08 & 19.51 & 44.37 & 68.78 \\
 & MoGe-3~\cite{kong2026moge3} & 36.73 & 48.23 & 92.87 & \second 45.72 & \second 62.18 & 87.23 & \second 67.89 & 80.80 & 90.09 & 33.79 & 67.14 & 70.83 \\
 & UniDepth-v2~\cite{piccinelli2025unidepthv2} & 41.44 & 52.57 & 91.70 & 40.42 & 56.51 & 87.11 & 66.89 & 79.36 & 88.69 & 20.26 & 41.66 & 68.74 \\
\midrule
\multirow{4}{*}{TAPIP3D (RGBD)$^{\dagger}$~\cite{tapip3d}} & DA3-Metric & 29.14 & 40.41 & 93.61 & 30.36 & 47.04 & 86.17 & 63.53 & 78.68 & 87.47 & 26.62 & 55.95 & 68.82 \\
 & DA3-Nested & \third 44.39 & \second 54.69 & 93.18 & 22.22 & 34.79 & 86.71 & 62.92 & 74.57 & 89.75 & 21.44 & 48.62 & 69.87 \\
 & MoGe-3 & 37.49 & 48.46 & 94.10 & 44.23 & \third 61.41 & 85.93 & 67.45 & \third 81.54 & 90.05 & \second 37.22 & \second 71.23 & \third 72.93 \\
 & UniDepth-v2 & 42.46 & 53.75 & 93.04 & 39.59 & 56.41 & 85.96 & 61.49 & 75.35 & 88.42 & 21.76 & 44.49 & 70.61 \\
\midrule
\multirow{4}{*}{Point4D$^{\ddagger}$~\cite{jeon2026point4d}} & DA3-Metric & 29.46 & 39.86 & 88.09 & 9.54 & 16.35 & 85.48 & 65.50 & 78.93 & 88.66 & 22.99 & 51.75 & 59.08 \\
 & DA3-Nested & 38.58 & 50.40 & 88.09 & 7.80 & 13.58 & 85.48 & 62.03 & 74.93 & 88.66 & 20.92 & 47.37 & 59.08 \\
 & MoGe-3 & 33.45 & 45.37 & 88.09 & 8.86 & 15.11 & 85.48 & 67.43 & 81.05 & 88.66 & 26.77 & 58.74 & 59.08 \\
 & UniDepth-v2 & 36.38 & 48.38 & 88.09 & 8.21 & 14.23 & 85.48 & 59.67 & 72.11 & 88.66 & 16.57 & 37.25 & 59.08 \\
\midrule
\multirow{4}{*}{AllTracker$^{\dagger}$~\cite{harley2025alltracker}} & DA3-Metric & 29.43 & 40.30 & \second 95.02 & 31.66 & 46.87 & \first 89.42 & 64.61 & 78.95 & \third 90.38 & 28.39 & 55.62 & 67.09 \\
 & DA3-Nested & \second 44.51 & 54.17 & \second 95.02 & 23.05 & 34.53 & \first 89.42 & 61.61 & 74.21 & \third 90.38 & 23.68 & 47.40 & 67.09 \\
 & MoGe-3 & 37.41 & 48.05 & \second 95.02 & \third 45.67 & 61.23 & \first 89.42 & 67.11 & 81.19 & \third 90.38 & \third 35.05 & \third 67.71 & 67.09 \\
 & UniDepth-v2 & 42.93 & 53.51 & \second 95.02 & 40.89 & 56.38 & \first 89.42 & 61.05 & 75.11 & \third 90.38 & 22.11 & 43.89 & 67.09 \\
\midrule
\multirow{4}{*}{CoTracker3$^{\dagger}$~\cite{karaev25cotracker3}} & DA3-Metric & 29.39 & 39.52 & \first 95.16 & 30.67 & 43.72 & \third 88.97 & 66.53 & 77.70 & \second 91.59 & 27.17 & 40.71 & \second 75.63 \\
 & DA3-Nested & 44.13 & 52.99 & \first 95.16 & 22.37 & 32.20 & \third 88.97 & 63.45 & 73.11 & \second 91.59 & 22.74 & 34.84 & \second 75.63 \\
 & MoGe-3 & 37.36 & 47.05 & \first 95.16 & 44.83 & 57.87 & \third 88.97 & \first 68.60 & 79.47 & \second 91.59 & 34.55 & 50.69 & \second 75.63 \\
 & UniDepth-v2 & 42.58 & 52.46 & \first 95.16 & 40.31 & 53.27 & \third 88.97 & 63.17 & 74.22 & \second 91.59 & 21.52 & 32.78 & \second 75.63 \\
\midrule
\multirow{4}{*}{4RC$^{\ddagger}$~\cite{luo20264rc}} & DA3-Metric & 24.96 & 37.80 & 79.04 & 21.49 & 35.54 & 79.70 & 62.37 & 80.88 & 82.49 & 12.01 & 30.70 & 40.02 \\
 & DA3-Nested & 29.37 & 42.73 & 79.04 & 16.79 & 28.10 & 79.70 & 58.62 & 76.35 & 82.49 & 11.29 & 28.19 & 40.02 \\
 & MoGe-3 & 28.62 & 43.02 & 79.04 & 21.89 & 35.04 & 79.70 & 64.50 & \first 83.47 & 82.49 & 13.85 & 33.75 & 40.02 \\
 & UniDepth-v2 & 27.76 & 40.35 & 79.04 & 18.85 & 30.29 & 79.70 & 59.08 & 76.66 & 82.49 & 8.09 & 20.15 & 40.02 \\
\midrule
\multirow{4}{*}{OmniX$^{\ddagger}$~\cite{jiang2026omnix}} & DA3-Metric & 24.77 & 36.28 & 82.91 & 7.44 & 12.90 & 83.28 & 50.90 & 67.74 & 83.12 & 18.53 & 45.54 & 49.20 \\
 & DA3-Nested & 34.89 & 48.44 & 82.91 & 6.02 & 10.63 & 83.28 & 47.34 & 63.60 & 83.12 & 17.88 & 43.66 & 49.20 \\
 & MoGe-3 & 29.28 & 42.74 & 82.91 & 6.98 & 12.03 & 83.28 & 54.79 & 73.26 & 83.12 & 22.15 & 53.67 & 49.20 \\
 & UniDepth-v2 & 30.21 & 42.56 & 82.91 & 6.08 & 10.66 & 83.28 & 53.72 & 71.61 & 83.12 & 14.00 & 34.96 & 49.20 \\
\midrule
\multirow{4}{*}{SpaTracker-v2$^{\ddagger*}$~\cite{xiao2025spatialtrackerv2}} & DA3-Metric & 32.58 & 42.12 & \third 94.86 & 36.77 & 53.10 & 88.80 & 64.32 & 77.61 & 89.33 & 26.33 & 55.17 & 72.48 \\
 & DA3-Nested & 43.44 & 52.75 & \third 94.86 & 24.04 & 35.64 & 88.80 & 60.38 & 72.90 & 89.33 & 23.74 & 49.81 & 72.48 \\
 & MoGe-3 & 36.29 & 47.20 & \third 94.86 & 43.56 & 60.71 & 88.80 & 66.96 & 81.31 & 89.33 & 31.33 & 64.90 & 72.48 \\
 & UniDepth-v2 & 41.55 & 51.17 & \third 94.86 & 38.95 & 54.58 & 88.80 & 65.22 & 79.03 & 89.33 & 20.09 & 42.84 & 72.48 \\
\midrule
\multirow{4}{*}{Any4D$^{\ddagger}$~\cite{karhade2025any4d}} & DA3-Metric & 25.08 & 35.75 & 89.03 & 30.77 & 46.49 & 83.60 & 56.71 & 73.38 & 84.62 & 9.63 & 25.98 & 45.70 \\
 & DA3-Nested & 27.83 & 38.03 & 89.03 & 22.80 & 34.47 & 83.60 & 53.35 & 68.75 & 84.62 & 9.07 & 24.86 & 45.70 \\
 & MoGe-3 & 26.87 & 38.64 & 89.03 & 32.32 & 46.86 & 83.60 & 60.66 & 78.60 & 84.62 & 11.12 & 29.27 & 45.70 \\
 & UniDepth-v2 & 30.08 & 40.52 & 89.03 & 29.12 & 42.17 & 83.60 & 60.23 & 78.33 & 84.62 & 6.32 & 16.79 & 45.70 \\
\midrule
 & Native
 & 40.87 & 51.59 & 93.76
 & 27.97 & 40.03 & 88.39
 & 34.80 & 43.63 & 89.28
 & 27.43 & 57.25 & 62.21 \\
\multirow{4}{*}{$\pi^{4}{}^{\ddagger}$~\cite{wang2026pi}} & DA3-Metric
 & 36.13 & 45.78 & 93.76 & 37.17 & 52.98 & 88.39
 & 61.91 & 74.73 & 89.28 & 26.94 & 57.69 & 62.21 \\
 & DA3-Nested & 43.91 & \third 54.46 & 93.76 & 25.00 & 36.39 & 88.39 & 58.50 & 70.73 & 89.28 & 23.15 & 50.09 & 62.21 \\
 & MoGe-3 & 39.27 & 50.14 & 93.76 & 42.88 & 59.54 & 88.39 & \third 67.50 & \second 82.10 & 89.28 & 31.11 & 65.49 & 62.21 \\
 & UniDepth-v2 & 40.73 & 51.13 & 93.76 & 33.63 & 49.10 & 88.39 & 63.20 & 77.86 & 89.28 & 20.44 & 44.85 & 62.21 \\
\midrule
\multirow{4}{*}{OpenD4RT$^{\ddagger}$~\cite{zhang2026d4rt,opend4rt}} & DA3-Metric & 19.76 & 29.12 & 90.81 & 7.62 & 15.61 & 77.21 & 61.10 & 73.81 & 90.00 & 22.38 & 46.98 & 67.19 \\
 & DA3-Nested & 22.79 & 32.49 & 90.81 & 6.20 & 13.18 & 77.21 & 57.35 & 69.42 & 90.00 & 19.86 & 42.49 & 67.19 \\
 & MoGe-3 & 22.73 & 33.02 & 90.81 & 7.59 & 15.62 & 77.21 & 62.01 & 75.50 & 90.00 & 25.02 & 52.88 & 67.19 \\
 & UniDepth-v2 & 21.84 & 31.59 & 90.81 & 6.88 & 14.56 & 77.21 & 59.16 & 72.05 & 90.00 & 18.35 & 39.41 & 67.19 \\
\cmidrule(l){1-14}
\textbf{Ours} (cam on) & None & \first 52.21 & \first 64.89 & 94.67 & \first 51.74 & \first 69.87 & \second 89.03 & 65.72 & 76.10 & \first 92.19 & \first 62.53 & \first 97.04 & \first 83.34 \\
\bottomrule
\end{tabular}%
}
\caption{\textbf{Additional depth providers for single-view metric tracking.}
Pi4 (native) retains its predicted scale without external depth.
DA3-Nested reference results from Table~\ref{tab:main_tracking} are repeated
for comparison; the embedded-DA3 reference is reported there. Our camera-on model receives known intrinsics
and metric poses through MCPE and uses no external depth. Symbols follow
Table~\ref{tab:main_tracking}. \ifmetareport Camera inputs are listed in
Table~\ref{tab:tracking_depth_inputs}. \else Camera interfaces are detailed in
Appendix~\ref{app:input_protocol}. \fi Shading ranks the
\colorbox{tabfirst}{best}, \colorbox{tabsecond}{second}, and
\colorbox{tabthird}{third} configurations shown here, with ties at displayed
precision sharing a rank.}
\label{tab:single_view_depth_variants}
\end{table*}

\section{Additional Qualitative Results}
\label{app:in_the_wild}
Fig.~\ref{fig:in_the_wild} shows eight more in-the-wild
recordings of people and robots, complementing
Figs.~\ref{fig:supp_ours} and~\ref{fig:supp_davis}.
Each example shows input frames of the query view and our
reconstruction with dense 3D trajectories, rendered from a
novel viewpoint.

\end{document}